\documentclass[final,3p,twocolumn]{elsarticle}

\usepackage{graphicx,subcaption}
\usepackage{comment}
\usepackage{amsmath}
\usepackage{hyperref}
\usepackage{xcolor}
\usepackage{soul}
\usepackage{ulem}
\usepackage{multicol, blindtext, multirow}
\usepackage{amssymb}

\usepackage{lineno}
\journal{Reliability Engineering $\&$ System Safety}

\begin{document}
\begin{sloppypar}
\begin{frontmatter}



\title{Domain Adaptation via Alignment of Operation Profile for Remaining  Useful Lifetime Prediction}


\author[inst1]{Ismail Nejjar}
\affiliation[inst1]{organization={Swiss Federal Institute of Technology Lausanne (EPFL)},
            country={Switzerland}}

\author[inst2]{Fabian Geissmann}
\author[inst1]{Mengjie Zhao}
\author[inst3]{Cees Taal}
\author[inst1]{Olga Fink}

\affiliation[inst2]{organization={Swiss Federal Institute of Technology Zürich (ETH Zürich)},
            country={Switzerland}}
\affiliation[inst3]{organization={SKF Group},
            country={Netherlands}}

\begin{abstract}

Effective Prognostics and Health \textbf{Management} (PHM) relies on accurate prediction of the Remaining Useful Life (RUL). Data-driven RUL prediction techniques rely heavily on the representativeness of the available time-to-failure trajectories. Therefore, these methods may not perform well when applied to data from new units of a fleet that follow different operating conditions than those they were trained on. This is also known as domain shifts. Domain adaptation (DA) methods aim to address the domain shift problem by extracting domain invariant features. However, DA methods do not distinguish between the different phases of operation, such as steady states or transient phases. This can result in misalignment due to under- or over-representation of different operation phases. This paper proposes two novel DA approaches for RUL prediction based on an adversarial domain adaptation framework that considers the different phases of the operation profiles separately. The proposed methodologies align the marginal distributions of each phase of the operation profile in the source domain with its counterpart in the target domain. The effectiveness of the proposed methods is evaluated using the New Commercial Modular Aero-Propulsion System (N-CMAPSS) dataset, where sub-fleets of turbofan engines operating in one of the three different flight classes (short, medium, and long) are treated as separate domains. The experimental results show that the proposed methods improve the accuracy of RUL predictions compared to current state-of-the-art DA methods.

\end{abstract}


\begin{keyword}
Remaining Useful Lifetime  \sep Deep Learning \sep Domain Adaptation \sep Prognostics
\PACS 0000 \sep 1111


\MSC 0000 \sep 1111
\end{keyword}

\end{frontmatter}


\section{Introduction}
\label{sec:Introduction}

 Remaining Useful Life (RUL) prediction, also referred to as prognostics, is a key element of Prognostics and Health Management (PHM). Prognostics aim to estimate how long a system can maintain its specified functionality before reaching its end of life \cite{si2011remaining}. Accurate RUL predictions enable scheduling maintenance from the perspective of operational and resource availability, avoid costly downtime, and prevent critical failures. In recent years, technological advancements and decreasing sensor costs have led to an increase in the amount of data collected from industrial assets \cite{zhao2019deep}. This increased data availability allows taking advantage of advanced data-driven approaches such as Deep Neural Networks (DNNs), which have recently shown great potential in PHM applications such as fault detection and diagnosis \cite{fink2020potential}. 

Traditional supervised machine learning methods assume that, on the one hand, a representative labeled dataset is available and, on the other hand, that the training and test datasets stem from similar distributions. Unfortunately, in real-world scenarios, both assumptions do not hold for many applications \cite{rombach2021contrastive,rombach2023controlled}. While these challenges are also present in fault detection and diagnostics, they are particularly pronounced in prognostics. On the one hand, collecting a sufficiently representative dataset of run-time failure trajectories is often impossible due to the potentially catastrophic consequences of such an event in reality. On the other hand, deep learning models still face the challenge of domain shift due to the wide variety of operating conditions and limited training samples. These two challenges considerably impact the generalization ability of a model trained on a specific dataset and applied to a different but related dataset.

Unsupervised Domain Adaptation (UDA) has been shown to be effective in addressing the domain gap problem by adapting a model that has been trained on one domain (the source) to a different, unlabeled domain (the target).
Previous research on domain adaptation has largely focused on classification tasks \cite{wilson2020survey}, and therefore, UDA has been extensively applied to fault diagnostics tasks \cite{li2020systematic,wang2019domain}. Recently, several research studies have also proposed applying UDA to prognostics tasks. These approaches typically involve aligning the feature distributions between the source and target domains through adversarial training \cite{da2020remaining}  or reducing the disparity between feature distributions \cite{cheng2021transferable, zhuang2021temporal}. Although adversarial training has been widely used in prognostics, it has not taken into account the distinct phases of cyclical operation profiles, such as those found in flights, for example, take-off, cruise and landing phases. A distinct marginal distribution defines each phase of the operation profile, and adversarial adaptation methods may fail to capture such multi-modal structures in condition monitoring data. Therefore, the alignment of the source and target domains using DANN~\cite{ganin2015unsupervised} assumes that the marginal distributions of each phase of the operation profile are also aligned. However, it may lead to misalignment of the phase of the operation profile due to under- or over-representation since the different operation phases may have different durations in the different operation profiles, for example.

To improve the ability to adapt between different domains, we propose utilizing the common sequence of operation profile phases which each system undergoes. We take advantage of the prior knowledge of the operating cycles since the phases always occur in the same order, share specific characteristics, and their existence and sequence are invariant across all domains.
For instance, in the aviation domain, a flight can be divided into several phases, that a flight follows, such as take-off, cruise, and descent.  The flight phases differ in terms of, e.g., altitude, duration, speed, etc., and have distinct marginal distributions. Within the same flight class, e.g., the cruise phase will have similar characteristics. However, between different flight classes, the characteristics of the flight phases will be more different. Each phase has a different impact on engine stress and degradation and consequently also on the RUL. By incorporating the operation profile into domain adaptation, we aim to improve the target RUL prediction in the case of a domain shift. 

In this paper, we consider the scenario where time-to-failure trajectories for complex systems such as aircraft jet engines and power plants are available within one fleet but not for units from a different fleet. To address this challenge, we define domains as sub-fleets of these systems that are operated under similar conditions. Our focus is on the domain adaptation problem, where we aim to transfer knowledge from a labeled source domain to an unlabeled target domain. The units within the same fleet may have a heterogeneous composition of operating conditions during their respective missions. These operating conditions can include differences in operating speed, machine load, working temperature, and environmental noise. These differences can lead to diverse distributions of marginal characteristics, making it difficult to adapt models to new domains. Previous research has shown that when the feature distribution is multimodal, adapting only the feature representation can be challenging for adversarial networks \cite{arjovsky2017towards,long2018conditional}.

To tackle this issue, we propose to replace the single-domain discriminator from DANN \cite{ganin2016domain} method with a discriminator for each phase of the operation profile. This approach is inspired by the success of multi-task learning \cite{zamir2018taskonomy}, and auxiliary tasks \cite{Wang_2021_ICCV} in facilitating adaptation between domains. The proposed methodologies align the marginal distributions of each phase of the operation profile in the source domain with its counterpart in the target domain. This work investigates in depth the learning of invariant features by adversarial learning for the alignment of sub-fleets (units operated in a similar way), taking into account their operation profiles. Two methods are proposed to deal with the above scenarios: 1) each part of the operational profiles can be assigned to a specific regime, or 2) the parts of the operational profile are smoothly assigned to all regimes with a defined probability, which is particularly useful for transition phases.

We evaluate the performance of the proposed algorithm on the NASA New Commercial Modular Aero-Propulsion System Simulation (N-CMAPSS) turbofan degradation dataset \cite{arias2021aircraft}. The dataset contains three flight classes: short (S), medium (M), and long (L), with a total of 15 units, with five units in each flight class. The proposed domain adaptation methods were evaluated on three adaptation tasks of increasing difficulty. The models were applied between sub-fleets, each consisting of five units, flying in a specific flight class. \textcolor{black}{The adaptation scenarios consist of transferring from medium to long flights ($M \rightarrow L$), from short to medium flights ($S \rightarrow M$), and finally from short to long flights ($S \rightarrow L$).} The adaptation scenarios always involve the adaptation from a flight class where run-to-failure trajectories are available to a flight class without any labels. The results of the three tasks demonstrate that the proposed algorithm, which takes into account the operation profile, can improve over the traditional DANN \cite{ganin2016domain} and outperform other comparative benchmark methods.

\section{Related Work}
\label{sec:Related Work}

\subsection{Unsupervised Domain Adaptation}\label{sec:UDA}

Unsupervised domain adaptation~(UDA)~\cite{pan2009survey} aims to improve the performance of a model on a target domain in the presence of a domain shift between the labeled source domain and unlabeled target domain. Several UDA methods have been proposed to align the feature distributions between the two domains during training using either discrepancy losses or adversarial training.

Prior works mainly relied on distribution alignment. While Deep Adaptation Networks~(DAN)~\cite{long2015learning} minimize the Maximum Mean Discrepancy (MMD) over domain-specific layers, Joint Adaptation networks \cite{long2017deep} align the joint distributions of domain-specific layers across different domains based on a Joint Maximum Mean Discrepancy (JMMD). In contrast,  adversarial-based methods strive to obtain domain-invariant representations through adversarial training. For instance, Domain-Adversarial Training of Neural Networks (DANN) \cite{ganin2016domain} aims to create a domain-invariant representation by using a domain discriminator. To achieve this, the model is trained with a reversal gradient layer to make the feature space indistinguishable for different domains. Maximum Classification Discrepancy (MCD) \cite{saito2018maximum} is another adversarial-based method that aims to reduce domain divergence and align the distribution of a target domain by considering task-specific decision boundaries through the use of task-specific classifiers in an adversarial training approach. Other approaches focused on the distribution of features for both domains across the batch normalization layers. Such approaches include Adaptive Batch Normalization (AdaBN) \cite{li2016revisiting} and Automatic Domain Alignment Layers
(AutoDial) ~\cite{carlucci2017autodial}, which aligns the distributions via modified batch normalization layers.

Most DA approaches have been developed for classification tasks, and domain adaptation for regression has received little attention. While classification approaches aim at generating decision boundaries to separate data into different classes, regression methods, on the other hand, aim at predicting continuous numerical outputs with a defined ordinal relationship.
Early works on domain adaptation for regression problems introduced a weighting scheme that assigns a weight according to the similarity between the training and test samples \cite{cortes2011domain,mansour2009domain}.
Recent works in the deep representation learning regime proposed different strategies to close the domain gap. 
For example, Representation Subspace Distance  (RSD) \cite{chen2021representation} explores the Riemannian geometry of the Grassmann manifold to reduce the domain gap using the orthogonal bases representation of the subspace as opposed to instance representations. \textcolor{black}{More recently, taking inspiration from the closed-form solution of least-square problems, DARE-GRAM \cite{nejjar2023dare} proposed a strategy centered on aligning the inverse Gram matrices of source and target features, with the aim of mitigating this issue.}

Many of the mentioned methods for classification can be naturally extended to regression problems. Nevertheless, when dealing with complicated regression problems, there are still no straightforward solutions to the fundamental problem of unsupervised domain adaptation for regression.

\subsection{Domain Adaptation applied to PHM} \label{sec:discuss_regression}

In recent years, domain adaptation has been increasingly applied to PHM applications \cite{li2022perspective}, particularly in the area of fault diagnostics classification. Three main domain gaps exist in the PHM context: between varying operating conditions \cite{zhang2020unsupervised,pan2019approach}, between different units of a fleet \cite{michau2021unsupervised}, and between simulations and real data \cite{GAO2021356,doi:10.1177/1475921720980718,YANG2019692,wang2021integrating}. 

Most research in this area has focused on the transfer between discrete operating conditions. Classic domain adaptation methods, such as distribution alignment \cite{lu2016deep,li2019multi} and adversarial alignment \cite{zhang2018adversarial,bao2021enhanced}, have been widely used in this context and have been found to be beneficial \cite{wang2019domain}. Proposed approaches include discrepancy-based domain adaptation, using various criteria such as Maximum Classification Discrepancy (MCD) \cite{bao2021enhanced} and Maximum Mean Discrepancy (MMD) \cite{8543590,han2021hybrid,9136845}. Deep domain adversarial frameworks such as DANN \cite{2020MSSP14506962J,aerospace9090516}, \textcolor{black}{\cite{liu2023intelligent}} have been developed and applied for fault diagnosis of machines under varying working conditions. Some works have sought to improve upon adversarial alignment through the use of conditional discriminators \cite{zhang2021conditional,yu2020conditional}.

\textcolor{black}{Remaining useful life predictions have received increasing attention. \cite{li2023remaining} proposes a novel RUL prediction approach that
combines knowledge of the sensor relationship and deep learning models.   
Recently, \cite{xiong2023adaptive} proposed an adaptive framework that can select the correct trained model according to failure modes. DSCN-DTAM was proposed in \cite{cheng2022two} and employs multiple regularization strategies and a double transferable attention mechanism to improve feature transferability in the presence of shift. Furthermore, transfer learning has also been recently increasingly applied to RUL prediction to increase the transferability of the learned feature,\cite{zhang2018transfer, ding2021remaining, cheng2022two}}. For example, a transfer learning algorithm based on Bidirectional Long Short-Term Memory (BLSTM) recurrent neural networks was proposed in \cite{zhang2018transfer} for RUL estimation. The algorithm fine-tunes a model trained on a large amount of data from a source task with a small amount of data from a target task, usually a different but related task. The results showed that transfer learning is effective, except when transferring from a dataset of multiple operating conditions to a dataset of a single operating condition, leading to negative transfer learning. In addition to cases where labels are available for the target, also unsupervised domain adaptation approaches have been applied for RUL prediction. Different adversarial methods have been proposed to align the source and target domains and improve the performance of RUL prediction \cite{da2020remaining}. For example, \cite{da2020remaining} used adversarial training to extract domain-invariant features from a Long Short-Term Memory (LSTM) model. Recent work \cite{hu2022remaining} introduced a novel Wasserstein distance-based weighted domain adversarial neural network (WD-WDANN) for RUL prediction under different operating conditions by measuring the similarity of source samples to the target domain to determine the sample quality.  Multiple-kernel maximum mean discrepancies (MK-MMD) were proposed and shown to be more robust than single-kernel methods \cite{zhu2020new,zhuang2021temporal} and help to minimize the distribution discrepancy between different failure behaviors in the feature space.

Previous approaches proposed for RUL prediction in the context of PHM have mainly focused on applying domain adaptation on the entire operating cycle without considering the distinct phases of the operation profile for each domain. Such an approach aligning source and target domains using domain invariant feature learning implicitly assumes that the marginal distributions of different operating conditions are also aligned. We propose replacing the single-domain discriminator with multiple-domain discriminators for each operating condition. In this way, we ensure that the different marginal distributions are aligned with their counterparts in the other domain.

\section{Methodology}
\label{sec:Methodology}

\subsection{Terminology}
\label{sec:terminology}

Before delving into the details of the proposed algorithm, it is important to define some key terms that will be used throughout the text. In this study, we make use of the terms \textit{operation profile} and \textit{operation condition}.

\noindent\textbf{An operation condition}: refers to the specific state or set of conditions under which a system is operating and can be thought of as different domains.

\noindent\textbf{An operation profile}: is a detailed and specific way of describing how industrial assets are operated and are controlled under different conditions while sharing similar characteristics in the sequences in which the operating conditions occur. An operation profile is typically composed of \textbf{distinct and discrete phases}, which are characterized by definite control and condition parameter ranges defining the states of the system that are retained for specific periods of time and are characterized by a unique marginal distribution of its characteristics. 

Together, the operation profile (with different phases) and operation conditions provide a comprehensive understanding of how industrial assets function and perform over time.

\subsection{General Problem Definition}\label{sec:problemdef}

In this paper, we study the unsupervised domain adaptation problem in the context of regression.  During training, we are given access to a set of $n_{s}$ labeled samples from a source domain $D_s = \{(x_{s}^{i}, y_{s}^{i})\}_{i=1}^{n_s}$ and $n_{t}$ unlabeled samples from a target domain $D_t = \{(x_{t}^{i})\}_{i=1}^{n_t}$. 

Each observation \textcolor{black}{$x^i \in \mathcal{X}\subset \mathbb{R}^{K\times T}$}
is a multivariate time sequence of \textcolor{black}{$K$} raw measurements of length $T$, where $T$ is the length of the observation window.
Finally, \textcolor{black}{given a sample $i$ from the set of all trajectories in the source domain,} $y^i\in \mathcal{Y} \subset \mathbb{R}$ denotes the Remaining Useful Lifetime (RUL) value of the observed sequence $x^i$. Labels of target data are not available during training. The goal of the task is to train a model using labeled set $D_s$ and unlabeled set $D_t$ multivariate time-series data of aircraft engines operated under different operating conditions.

\subsection{Operation profile}\label{sec:Operating Conditions}

\begin{figure}[ht]
    \centering
    \begin{subfigure}[t]{0.38\linewidth}        
        \centering
        \includegraphics[width=\linewidth]{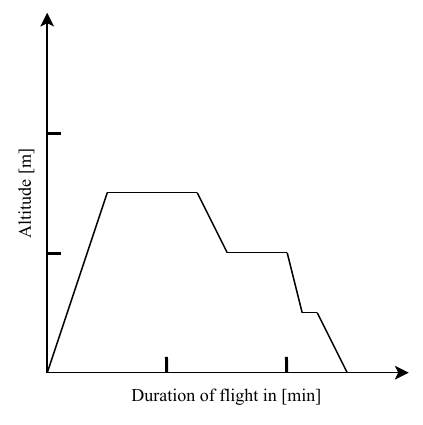}
        \caption{Operating Condition: Short flight Altitude profile}
        \label{fig:A}
    \end{subfigure}
    \begin{subfigure}[t]{0.5\linewidth}        
        \centering
        \includegraphics[width=\linewidth]{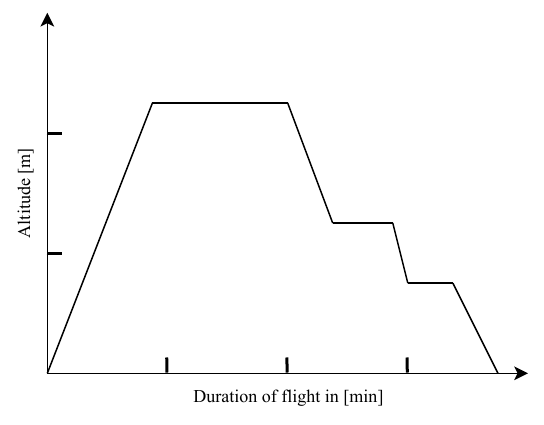}
        \caption{Operating Condition: Long flight Altitude profile}
        \label{fig:B}
    \end{subfigure}\\
    \begin{subfigure}[b]{0.7\linewidth}        
        \centering
        \includegraphics[width=\linewidth]{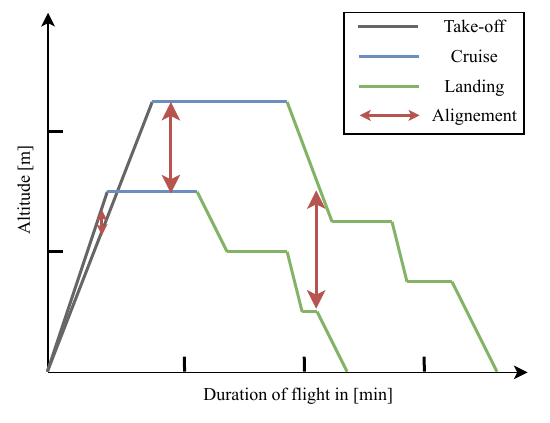}
        \caption{Altitude level at the different phases of the flight Take-off, cruise, and Landing.}
        \label{fig:C}
    \end{subfigure}
    \caption{The figure \ref{fig:A} and \ref{fig:B} show two flights that were operated differently and have different profiles. Figure  \ref{fig:C} illustrates the presence of similar phases in both flights, which were used for the alignment process.}
    \label{fig:profile}
\end{figure}

Industrial assets may operate under different operating conditions but share similar phases of the operation profile. For example, in aviation, a flight is usually composed of different phases, such as take-off, cruise, and descent, requiring different thrust levels to propel the aircraft forward (Fig.\ref{fig:profile}), thus applying different types and levels of stress on the jet engine.   

More generally, the operation profile can be determined by discretizing a particular measured or control parameter based on domain knowledge or using an unsupervised clustering algorithm on the control parameters or multivariate observations. The final objective is to obtain a set of $n_p$ discrete and common phases of the operation profile for each of the two domains. 

This research assumes that a finite number of discrete operation profile labels can either be provided based on domain knowledge or identified using the measured or control parameters. We argue that taking into account the different phases of the operation profile can improve the alignment process.

The following section presents how these operation profile labels can be exploited to extend upon the existing DANN methods to support the alignment process. 

\subsection{Domain-Adversarial Neural Networks}

As the operating conditions of the source and target sub-fleets differ, a model trained on the source data would be able to easily distinguish the feature vectors of the target domain from the source domain due to the domain shift, leading to poor performance when applied to the target domain. Adversarial distribution alignment methods aim to address this problem by ensuring that the feature extractor is unbiased with respect to the characteristics of the source and target domains.

DANN \cite{ganin2015unsupervised} is a broadly applied domain adaptation method that aligns the distributions of source and target features by adding a domain discriminator and introducing adversarial training. 

\begin{figure}[ht]
    \centering
    \includegraphics[width=1\linewidth]{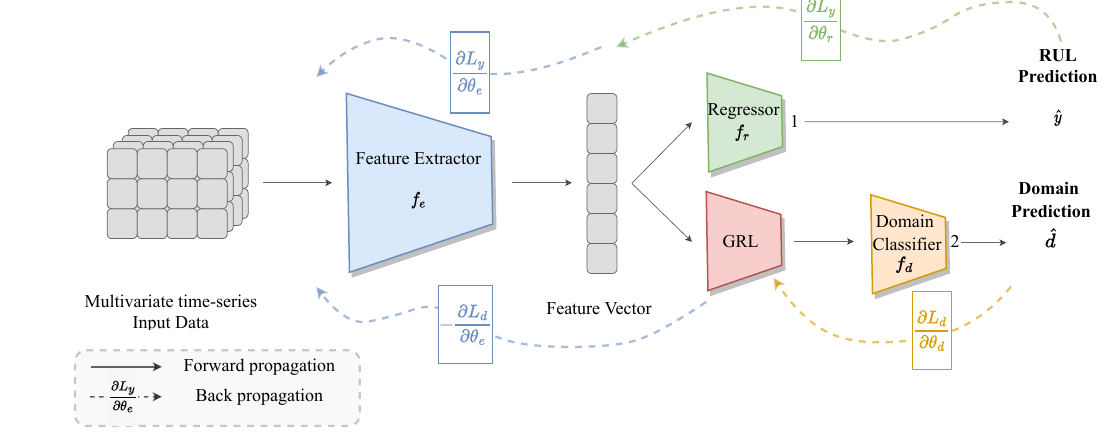}
    \caption[DANN)]{Standard architecture of DANN approaches. The gradient Reversal Layer (GLR) ensures that the feature distributions over the two domains are  indistinguishable for the domain classifier}
    \label{fig:dann}
\end{figure}

As illustrated in Figure \ref{fig:dann}, the DANN model is composed of a feature extractor $f_e$, a regressor $f_r$, and a domain classifier $f_d$ parameterized by $\theta_{e},\theta_{r},\theta_{d}$ respectively.

The feature extractor $f_e$ takes the input data $x$ and learns a feature representation denoted by $f_e(x)$. 

The regressor predicts the RUL label for each input sample $\hat{y} = f_r(f_e(x))$. The RMSE loss is used for minimizing the error between the true and the predicted RUL on the source data (for the RUL prediction task):

\begin{figure*}[ht]
    \centering
    \includegraphics[width=1\linewidth]{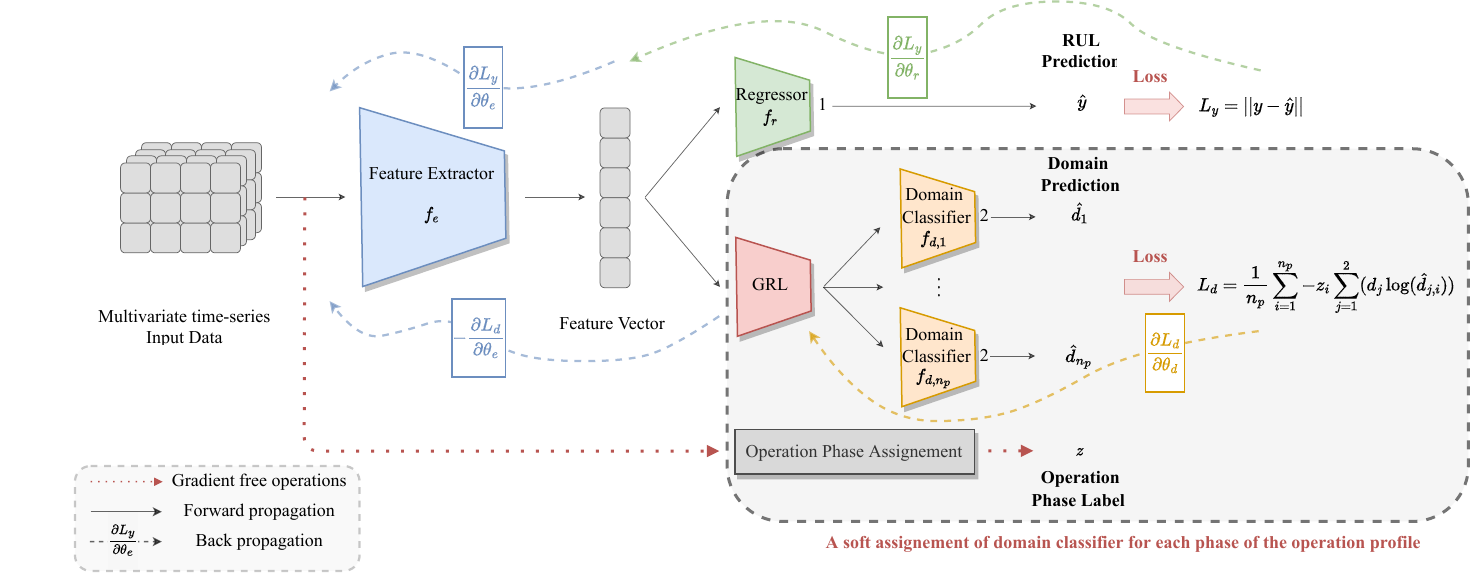}
    \caption[OPS-DANN (hard)]{Proposed OPS-DANN (hard) approach. The source and target samples of each operation profile are aligned with a dedicated domain discriminator.}
    \label{fig:ocs_dann_hard}
\end{figure*}

\begin{equation}
    \mathcal{L}_{y,s} (\theta_{e}, \theta_{r} ) =  \sqrt{(\hat{y}_s - y_s)^{2}}
    \label{eq:loss_rul}
\end{equation}

The second part of the model includes a domain discriminator, which is responsible for aligning the two domains. The domain classifier predicts the domain label for each input sample $\hat{d} = f_d(f_e(x))$, where $d \in \{0, 1\}$ is a domain label assigned to each training example to indicate its origin. The domain classifier uses the binary cross entropy loss function shown in Equation \ref{eq:loss_bce}.

\begin{equation}
    \mathcal{L}_{d} (\theta_{e}, \theta_{d}) = -\: \big(d \: \text{log}(\hat{d}) + (1 - d) \: \text{log}(1 - \hat{d})\big)
    \label{eq:loss_bce}
\end{equation}

While the domain discriminator weights $\theta_{d}$ are trained to minimize the binary cross-entropy loss, the feature extractor weights $\theta_{e}$ are updated to maximize the binary cross-entropy loss. The feature extractor learns to extract domain-invariant features, rendering the domain discriminators incapable of predicting the true domain label. These two contradicting objectives are trained in an adversarial procedure utilizing a gradient reversal layer (GRL).
GRL is an operation that acts as an identity function during the forward pass and reverses the sign of the gradient during the backward pass. The GRL allows us to implement this optimization problem in practice easily \cite{ganin2015unsupervised}.

\textcolor{black}{Finally, the regular stochastic gradient solvers (SGD) can be adjusted to optimize the model as suggested by Equations \ref{eq:minmax1} and \ref{eq:minmax2}.}

\begin{equation}
   (\hat{\theta}_{e}, \hat{\theta}_{r}) = \text{arg} \: \underset{\theta_{e}, \theta_{r}}{\text{min}} \mathcal{L}(\theta_{e}, \theta_{r}, \hat{\theta}_{d})
   \label{eq:minmax1}
\end{equation}

\begin{equation}
    \hat{\theta}_{d} = \text{arg} \: \underset{\theta_{d}}{\text{max}} \mathcal{L}(\hat{\theta}_{e}, \hat{\theta}_{r}, \theta_{d})
    \label{eq:minmax2}
\end{equation}

\subsection{Operation profile-specific Alignment}\label{sec:ocs}

In this research, we propose the Operation Profile-Specific (OPS) alignment framework, which aims to align the marginal distributions of the specific phases of the operation profile between different domains. These phases are often overlooked by other domain-invariant feature-learning techniques. Our approach extends the alignment component of DANN to consider each operating phase individually when aligning the source and target domains. In previous research, it has been demonstrated that when the feature distribution is multimodal, adapting only the feature representation can be challenging for adversarial networks \cite{arjovsky2017towards,long2018conditional}. Therefore, to improve DANN's performance in these scenarios, we chose to use the DANN method, as it has a strong track record in domain adaptation tasks and has previously been successful in regression tasks for PHM \cite{da2020remaining}. We propose two different approaches for operation profile-specific alignment:

\begin{itemize}
    \item Hard \textcolor{black}{assignment} : OPS-DANN (hard)
    \item Soft \textcolor{black}{assignment}: OPS-DANN (soft)
\end{itemize}

\textcolor{black}{
For hard assignment, each sample is strictly assigned to its corresponding phase within the operational profile. In contrast, soft assignment involves assigning samples to multiple discriminators, which can be advantageous when operation phases are uncertain. This approach proves particularly useful during transitional periods between phases, where samples could be associated with more than one operation phase.}

\begin{figure*}[ht]
    \centering
    \hspace*{-1cm}%
    \includegraphics[width=1\linewidth]{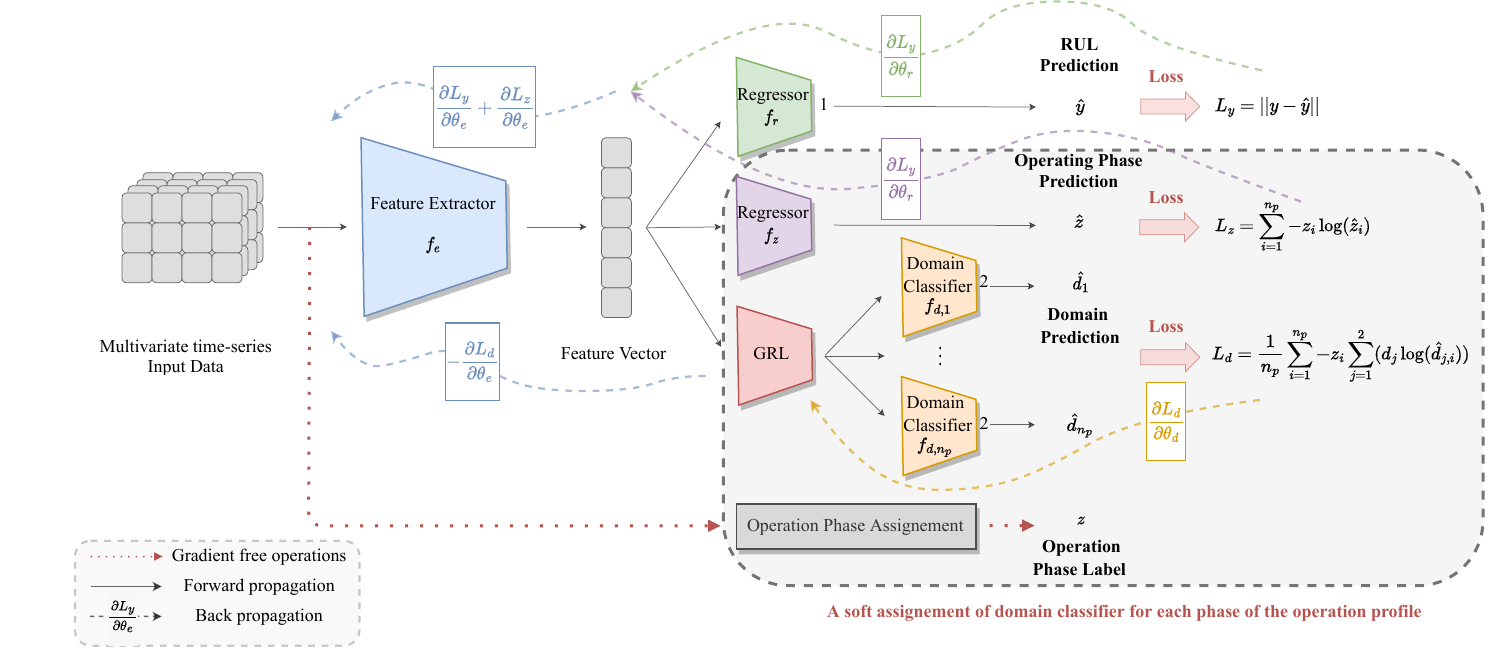}
    \caption[OPS-DANN (soft)]{Proposed  OPS-DANN (soft) approach. An additional operation profile classifier is added to allow for the assignment of each sample to multiple domain discriminators depending on its probability of belonging to the respective operating regime.}
    \label{fig:ocs_dann_soft}
\end{figure*}

\subsubsection{OPS-DANN (hard)}

Our first proposed extension to implement OPS alignment involves using dedicated domain discriminators for each operation profile. A hard assignment is made for each sample to its respective phase of the operation profile. The goal of OPS alignment is to separately align the distinct marginal distributions of each regime of the operation profile across the two domains.

This can be achieved by replacing the single domain discriminator task that discriminates between the source and the target domain with as many individual discriminators as there are discrete operating phases. Each individual domain discriminator still aims to differentiate between the source and the target domain. However, it only does so for samples belonging to one operating phase. This ensures that the marginal distribution of each operating phase is aligned with its counterpart in the other domain. The proposed framework is visualized in Figure \ref{fig:ocs_dann_hard}.

The domain discriminators are separated for each $n_p$ operating phases. Each domain discriminator is parameterized with its own set of weights $\theta_{d_{j}}$ with $j \in \{1,\dots,n_p\}$ and aims to predict the domain label $\hat{d}_{j, i} = f_{d_{j}}(f_e(x))$ for a sample that is assigned to the $i$-th domain discriminator. The operating phase label $z^{i}$ assigns each sample to one of the $n_p$ discriminators. The binary cross entropy loss function shown in Equation \ref{eq:loss_bce_hard} is utilized to determine the loss of each domain discriminator, where $z^{i}$  is a binary variable indicating whether a sample belongs to operating phase $i$ or zero otherwise. The new domain discriminator loss is presented in Equation \ref{eq:loss_bce_hard}:

\begin{equation}
    \mathcal{L}_{d,i} (\theta_{e}, \theta_{d}) = -z_i\: \big(d \: \text{log}(\hat{d}) + (1 - d) \: \text{log}(1 - \hat{d})\big)
    \label{eq:loss_bce_hard}
\end{equation}

\subsubsection{OPS-DANN (soft)}

A soft assignment of samples to multiple discriminators can be useful when the operation phase labels are not certain. This can happen, for example, when a system is transitioning between two phases, and samples can be considered to belong to both operation phases.
Instead of using a hard assignment of each sample to a single phase of the operation profile, we propose using a probabilistic assignment (soft assignment). To achieve this, we add an additional classifier after the feature extractor to classify the operation phase. This classifier is trained in a supervised manner using the available operating phase labels and outputs a probability distribution over the $n_p$ phases. The soft assignment allows samples to be assigned to multiple domain discriminators when the operating phase classifier is uncertain. The predicted probabilities are then used to weigh each sample's contribution to each of the $n_p$ domain discriminators. This model is depicted in Figure \ref{fig:ocs_dann_soft}.

For OPS-DANN (soft), the previously introduced equations change slightly due to the additional operating phase classifier parameterized by $\theta_{z}$. For each source and target sample, it predicts a probability of a sample belonging to each of the $n_p$ operating phases $\hat{z} = f_z(f_e(x))$. The prediction $\hat{z}$ can be written as a $n_p$-dimensional vector $\hat{z} = [\hat{z}_{1}, \dots, \hat{z}_{n_p}]$. The cross-entropy loss function compares the model's predictions to the true operating phase labels.
\begin{equation}
    \mathcal{L}_{z}^{i} (\theta_{f}, \theta_{z}) = - z^{i} \: \text{log}(\hat{z}^{i})
    \label{eq:loss_ce}
\end{equation}

The final loss for the OPS-DANN (soft) can be written as : 
\begin{equation}
\mathcal{L}_{d,i} (\theta_{e}, \theta_{d}) = -\hat{z}_i\: \big(d \: \text{log}(\hat{d}) + (1 - d) \: \text{log}(1 - \hat{d})\big)
    \label{eq:loss_bce_soft}
\end{equation}

\section{Case Study}
\label{sec:Case Study}
 
\subsection{N-CMAPSS Dataset}
\label{sec:cmapss}
 
We evaluate our proposed methods for Domain adaptation on the new Commercial Modular Aero-Propulsion System Simulation~(N-CMAPSS) dataset, which contains run-to-failure trajectories of large turbofan engines \cite{arias2021aircraft}. We compare the performance to previously proposed DA approaches.  The dataset was created using NASA's high-fidelity simulation model \cite{frederick2007user}, which allows the simulation of flight data over a wide range of flight conditions. Concretely, the flight data covers take-off, cruise, and descend flight conditions corresponding to different commercial flight routes. Each engine unit is assigned to one of the three flight classes based on the duration of the individual flights. Short flights are defined as flights that last between one and three hours, medium flights are between three and five hours long, and long flights last longer than five hours. 

\begin{figure}[ht]
    \centering
    \includegraphics[width=0.9\linewidth]{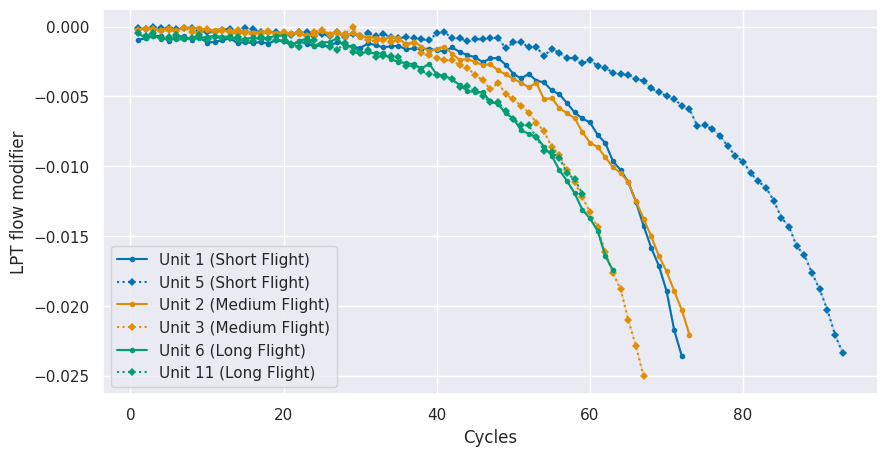}
    \caption[Degradation of Engine Units]{Example of degradation trajectories of engine units over time for engines from different domains. Vertical lines denote the fault initialization of each engine unit.}
    \label{fig:degradation}
\end{figure}

The degradation behavior of each engine unit is modeled as the combination of three contributors:
an initial degradation, a normal degradation, and an abnormal degradation due to a fault.  In the first phase, the engine degrades due to the normal degradation until the onset of a fault, whereby different fault types can occur. \textit{Normal degradation} is modeled linearly. Once a fault is initiated, the engine enters an \textit{abnormal degradation} phase until it ultimately reaches its end-of-life (EOL). During the abnormal degradation phase, the engine health decays exponentially. Figure \ref{fig:degradation} shows sample degradation trajectories of six engine units. 

The degradation trajectories are given as a multivariate time series of the sensor readings, containing measurements from 14 sensors from the turbofan engine, as well as scenario descriptors containing 4 parameters characterizing the operating condition of the flight, also referred to as scenario descriptors. The signals and the scenario descriptors are sampled at a frequency of 1 Hz. The different sensors include various temperature and pressure measurements, fan speeds, and fuel flow. The scenario descriptors include the altitude, speed, throttle-resolved angle, and total temperature at the fan inlet.

Each turbofan engine consists of five main components: the fan, the low and high-pressure compressor, and the low-and high-pressure turbine. During the simulated lifetime of each unit, one or a combination of several sub-components fails, leading to the onset of one of seven possible fault types. Fault severity increases over time. The N-CMAPSS dataset introduced several improvements compared to the popular original CMAPSS dataset \cite{saxena2008damage}, frequently used by researchers working on RUL-prediction tasks~\cite{li2018remaining, da2020remaining, chao2022fusing, fan2020transfer}. The N-CMAPSS differs with respect to two main aspects. First, it considers actual flight conditions recorded on board of a commercial aircraft. Secondly, it extends the modeling of degradation by linking the degradation process to its operational history. The RUL prediction aims to indicate how many flights (i.e., cycles) of a particular engine (unit) are left before its EOL.

The N-CMAPSS dataset \cite{arias2021aircraft} contains eight datasets, each with different fault types and flight classes.  In this study, we consider the different flight classes as distinct domains and focus on dataset three (DS03) of the N-CMAPSS dataset, which is characterized by a single fault type (HPT-efficiency failure combined with LPT flow and efficiency failure) and equally distributed flight classes. We define the sub-fleets operated under different flight classes as a domain. The subset used for this paper contains an equal number of five units per flight class. Despite the identical number of units, the total number of samples of the three domains varies considerably due to the different lengths of the flights, as shown in Table~\ref{tab:dataset_size}. 

\begin{table}
\centering
\begin{tabular}{cccc}
\hline
\textbf{Domain} & \textbf{Short} & \textbf{Medium} & \textbf{Long} \\
\textbf{Flights} & \textbf{(S)}&  \textbf{(M)}& \textbf{(L)}\\
\hline
Number of units      & 5                      & 5                       & 5                     \\
Number of samples            & 104k                   & 202k                    & 362k                  \\ \hline
\end{tabular}
\caption[Domains]{Number of units and samples of the three considered domains.}
\label{tab:dataset_size}
\end{table}

\begin{figure}[ht]
    \centering
    \includegraphics[width=0.95\linewidth]{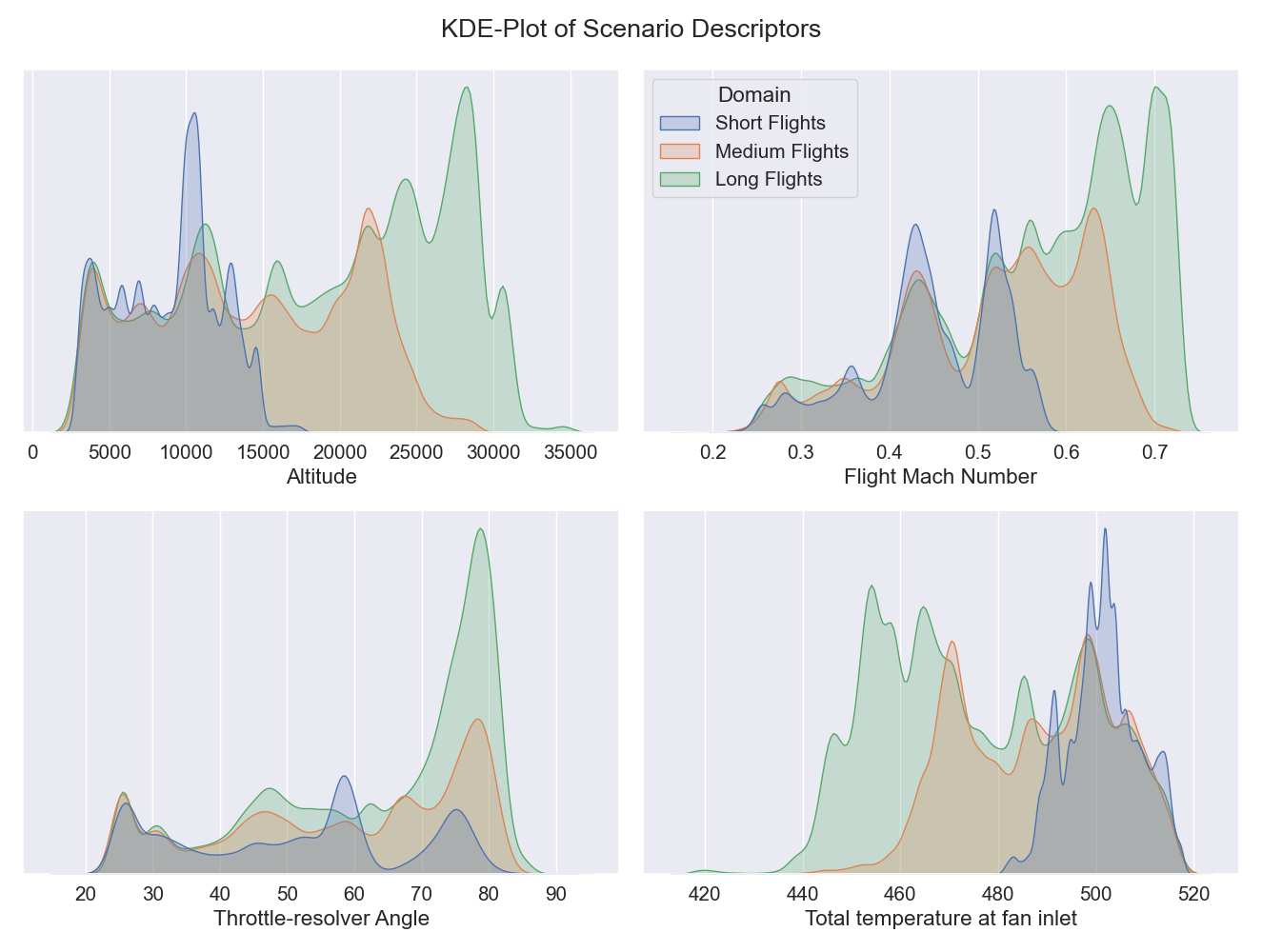}
    \caption[KDE-Plot of different Domains]{The kernel density estimate (KDE) shows the probability distribution of the four scenario descriptors for each domain separately. In three out of four cases, it can be seen that the long-haul flights span the widest range of feature values, while the short-haul flights cover a noticeably smaller part.}
    \label{fig:kdeplot}
\end{figure}

In aviation, engines are subject to high stress during take-off. Therefore, in this dataset, short-haul flights have more take-offs and landings compared to long-haul flights, resulting in a different type of degradation. In addition, the sensor measurements of long-haul flights are significantly different from those of short-haul flights, as they reach higher altitudes and faster speeds, as can be seen in Figure ~\ref{fig:kdeplot}. 

Given the three domains, there are six possible domain adaptation tasks, of which only the three presumably most difficult tasks are considered in this research. The three tasks comprise the transfer from a sub-fleet of short-range flights to the sub-fleets of mid-range and long-range flights, as well as from mid-range to long-range flights. These three tasks are particularly difficult because the marginal distribution of the characteristics of shorter flights does not cover those of longer flights, as visualized using the four scenario descriptors in Figure~\ref{fig:kdeplot}. The corresponding considered adaptation tasks are abbreviated as $S \rightarrow M$ for short to medium flight length adaptation tasks by $S \rightarrow L$ and $M \rightarrow L$ for the other two tasks, respectively.

\begin{figure}[ht]
    \centering
    \includegraphics[width=1.0\linewidth]{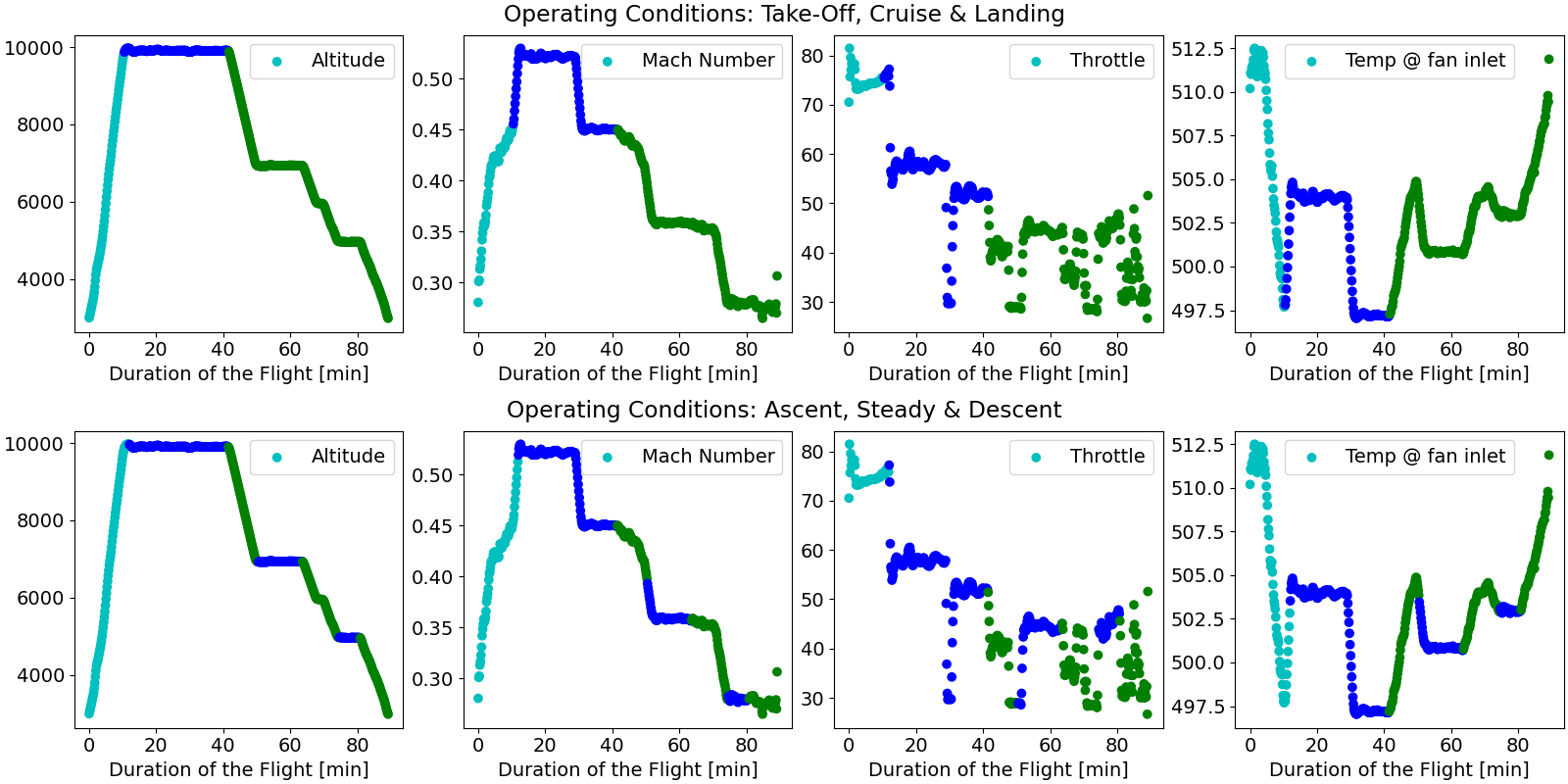}
    \caption[Operating Conditions]{Direct comparison of two possible separations of the operating conditions. Ascent and take-off are colored in turquoise, respectively; cruise and steady flight in dark black; and descent and landing in green.  The short periods of steady flying are differentiated from the rest using the change in altitude. }
    \label{fig:definition_oc}
\end{figure}

In general, it seems intuitive to subdivide a flight into take-off, cruise, and landing operating conditions. However, in this work, we found a better suitable division of the operation profile: in ascending, steady, and descending operating conditions. This division allows a more fine-grained separation and is more closely related to how the airplane is operated.
For example, the operating conditions of an airplane that changes its flight altitude during a flight would only be assigned to the cruise condition with the more intuitive partitioning approach. In contrast, the partitioning proposed in this research allows separating the two steady flight sequences from the descending one in between (see Figure~\ref{fig:definition_oc}).

Ascending, steady, and descending flight conditions can be identified using the first-order derivative of the altitude measurement. To that end, the change in altitude between neighboring sampling points was considered and grouped into the three operating conditions using a threshold value. The threshold was experimentally set to $T = 0.5 \frac{ft}{s}$. The Operation Phase label z can be defined based on the following inequalities: 
\begin{equation}
         \textrm{z} =  \left\{ 
  \begin{array}{ c l }
    \textrm{Ascending} & \quad \textrm{if }  \frac{x_{alt}^{i+1} - x_{alt}^i}{\Delta t} \geq T \\
    \textrm{Descending}           & \quad \frac{x_{alt}^{i+1} - x_{alt}^i}{\Delta t} \leq -T \\
    \textrm{Steady}           & \quad \textrm{otherwise}  
  \end{array}
\right.
\end{equation}

A median filter with a length of 51 was applied to smoothen the predictions.

\begin{figure}[ht]
    \centering
    \includegraphics[width=0.8\linewidth]{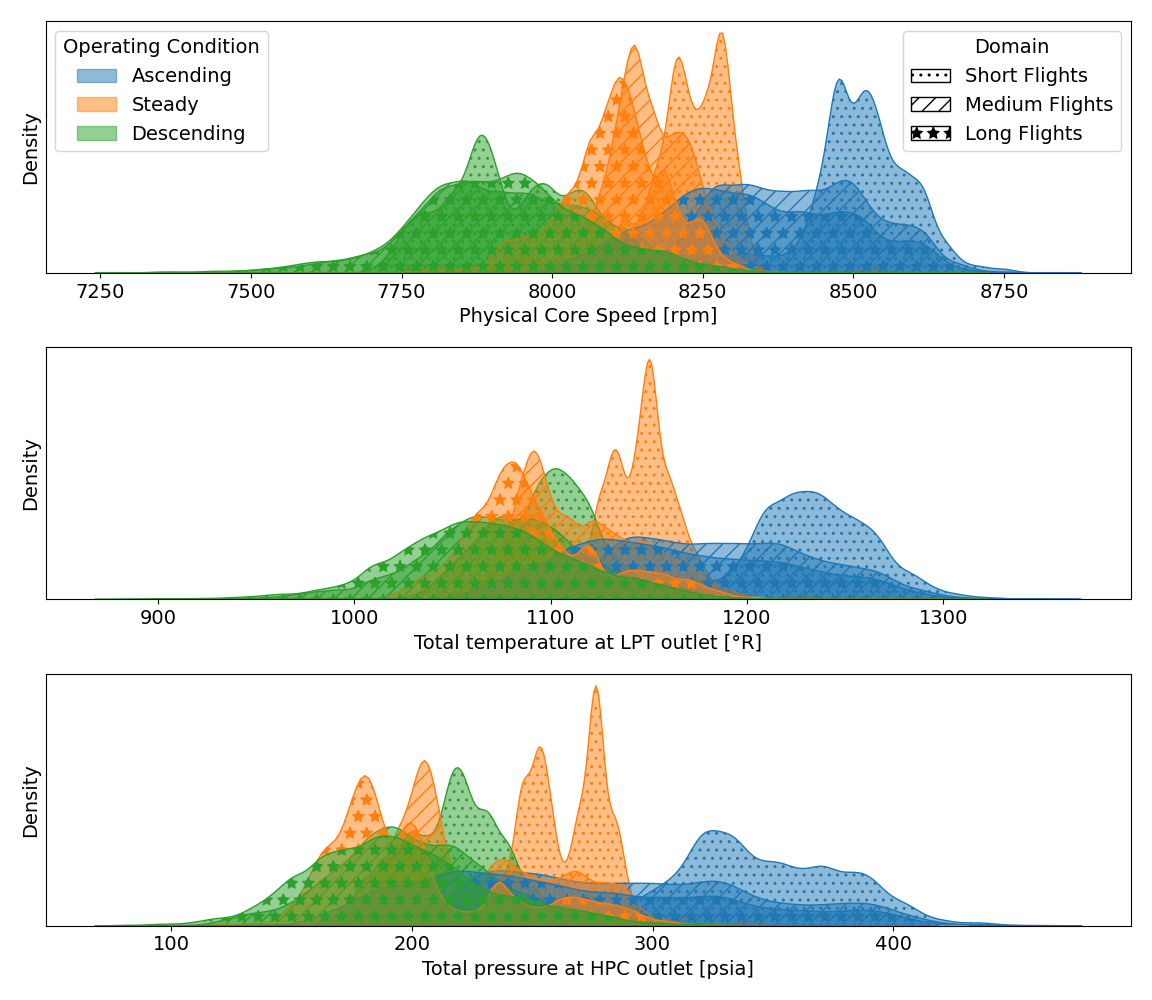}
    \caption[Marginal Distributions per Operating Condition]{Kernel density estimate for marginal distributions of each operation profile for the three domains. Three exemplary sensor values were selected: Physical Core speed, Total Temperature at the LPT outlet, and total pressure at the HPC outlet.}
    \label{fig:operating_conditions}
\end{figure}

The key idea behind the proposed OPS alignment is to consider each operation phase separately during the alignment to ensure that its marginal distribution is matched with its counterpart in the other domain. Figure \ref{fig:operating_conditions} shows the marginal distributions of three exemplary sensors for the short, medium, and long flight domains. Most of the other sensor values follow similar patterns. 

\subsection{Preprocessing}\label{sec:preprocessing}

Before a fault initiates, the normal degradation process is slow, and it can be assumed that the RUL is very large. Thus, for this work, we only start predicting the RUL after the onset of the fault. This results in a two-step process in which fault detection is performed first, and the RUL prediction is only initiated after the fault has been detected. In this paper, we are only focusing on the second step and assume that the fault onset detection is given. Therefore, the RUL prediction task aims to predict the remaining cycles after a fault has been initiated and detected.

Sensor measurements and scenario descriptors are taken as inputs to ensure a realistic usage scenario, resulting in 18 sensor values. Other quantities, such as virtual sensors and model health parameters, which require internal parameters of a simulator, are not considered in this research. 

The dataset was downsampled by a factor of ten using a Chebyshev filter with order eight, which applies an anti-aliasing filter before the down-sampling process. This ensured the same 0.1 Hz sampling frequency that was used in \cite{chao2022fusing}.

Each signal measurement was scaled to a range $[-1,1]$ using min-max normalization. The scaler was fitted on the source domain data for each adaptation task and subsequently applied to the target domain data.

The RUL was normalized for each engine unit to decrease from one to zero \textbf{between the point in time when the fault occurs and the end of life}. To that end, all samples belonging to one engine unit were divided by the maximum number of cycles of the respective unit.

\subsection{Model Architecture}\label{sec:modelarchitecture}

 In this research, we compare the proposed methodology to other commonly used DA approaches. All of the DA methods use the same basic architecture to ensure a fair comparison. In particular, we used a one-dimensional-convolutional neural network (1DCNN) with the architecture inspired by that proposed in \cite{chao2022fusing}.
The feature extractor consists of three 1D-convolutional layers (L=3). The first two layers have ten channels each, while the third layer condenses the feature representation into a single channel. Filters of size ten are used along with a stride of one, and zero-padding is added to ensure that inputs keep the same length when passing through the network. The network uses $ReLU$ as the activation function. The feature extractor has a total of 3k trainable model parameters.

The RUL-regressor contains two fully connected layers (L=2). The first one takes the flattened output from the feature extractor and passes it through a fully-connected layer with 50 neurons. Then, after another $ReLU$ activation function, the last layer predicts a single output value, the RUL, which is subsequently normalized to a range between 0 and 1 using a $Sigmoid$ activation function. In total, there are approximately 3k parameters in the RUL-regressor.

The domain discriminator has a similar architecture as the RUL regressor but contains an additional layer. A first fully-connected layer with 50 neurons is followed by a second one with 30 neurons. The last layer is also fully-connected and ends with a single output neuron. Again, a $Sigmoid$ activation function is applied to the final output, and $ReLU$ activation functions are used in between fully connected layers. The number of layers and neurons is loosely inspired by an earlier work, which also used a DANN architecture on a similar dataset \cite{da2020remaining}. The resulting domain discriminator has roughly 4k parameters.

\textbf{OPS-DANN (hard)}: This model utilizes three domain discriminators, as there are three different phases for the operation profile in the dataset used in this work. The architecture of each discriminator is identical to the one described above, leading to a total number of 18k parameters.

\textbf{OPS-DANN (soft)}: Similarly to the OCS-DANN (hard) version, this model utilizes three distinct domain discriminators. Additionally, it uses an operating phase classifier, which is added after the feature extractor. This classifier uses the same architecture as the domain discriminator, except for the last activation function. A $Softmax$ activation function is used in the model instead of a $Sigmoid$ activation function, and this is because there are multiple classes that the model needs to predict and not just two classes since $Softmax$ activation is designed to handle multi-class classification. OPS-DANN (soft) has the largest number of trainable model parameters, with a total of 22k.

\textbf{Multi-Class OPS-DANN}: We propose to explore the impact of the domain alignment of the operation phases in an ablation analysis. Compared to the architecture introduced above, this model only requires changes in the architecture of the domain discriminator of DANN. The domain discriminator predicts the domain of each sample and its operating phases. As there are three operating conditions, the discriminator requires six output neurons, one for each domain and operating phase pair. Consequently, the $Sigmoid$ function must be exchanged by the $Softmax$ activation to handle the multi-class output. This model has slightly more parameters than the original DANN, with roughly 10k parameters.

\subsection{Comparison Methods}\label{sec:baselines}

To fairly evaluate the performance of our proposed methods, we first establish a baseline using a feature extractor with a regressor that is trained only on the source data. We then compare the results of our Operation Profile-specific Domain Adaptation Network (OPS-DANN) and its variants to established domain adaptation techniques such as AdaBN, MK-MMD, and DANN. These methods have been previously applied to similar tasks in the field of prognostics and health management and have been shown to have strong performance, as reported in previous research studies such as \cite{wang2019domain}.
%
\textbf{MK-MMD}: This DA method is a domain-invariant feature learning technique. However, instead of using adversarial training, it aims to minimize a divergence measure between the two domains. To that end, the feature representation $f$, found by a feature extractor, is used to compute the MMD measure as shown in Equation \ref{eq:mmd}. Multiplications of the feature transformation $\phi(\cdot)$ can be readily computed by taking advantage of the kernel trick as demonstrated in Equation \ref{eq:kerneltrick}. For the multiple kernel version of MMD, $K$ Gaussian kernels with different bandwidth parameters $\gamma$ are combined to enhance model performance by adding them together, as shown in Equation \ref{eq:multikernel}.

\begin{equation}
    MMD(F_{S}, F_{T}) = \bigg\Vert \frac{1}{n_{S}} \sum_{i=1}^{n_{S}} \phi(f_{S}^{i}) - \frac{1}{n_{T}} \sum_{j = 1}^{n_{T}} \phi(f_{T}^{j})\bigg\Vert^{2}
    \label{eq:mmd}
\end{equation}

\begin{equation}
    k(f_{S}^{i}, f_{T}^{j}) = \langle \phi(f_{S}^{i}), \phi(f_{T}^{j}) \rangle
    \label{eq:kerneltrick}
\end{equation}

\begin{equation}
    k(f_{S}^{i}, f_{T}^{j}) = \sum_{k = 1}^{K} k_{k}(f_{S}^{i}, f_{T}^{j})
    \label{eq:multikernel}
\end{equation}

For this work, five kernels were selected, similar to previous applications as reported in\cite{cheng2021transferable}. The bandwidths parameter was set to 0.01, 0.1, 1, 10, and 100. 

\textbf{AdaBN}: Unlike  the other baseline methods introduced above, AdaBN \cite{DBLP:journals/corr/LiWSLH16} is not a domain-invariant feature learning technique. Instead, it aims to replace the normalization statistics of batch-norm layers computed on the source domain with those of the target domain. Notably, the target domain data is only used to update the normalization statistics, while all the other model parameters are trained using source data only. A necessary condition to utilize AdaBN is an architecture with batch-norm layers; consequently, the feature extractor's standard architecture had to be adapted. A 1D-batch-norm layer was placed following each of the three convolutional layers.

In summary, except for AdaBN, which uses additional batch-norm layers, all other baseline methods use the same architecture for the feature extractor and the RUL regressor to ensure a fair comparison.

\subsection{Training Procedure}\label{sec:trainingprocedure}

\begin{table*}
\small
\centering
\resizebox{1\textwidth}{!}{
\begin{tabular}{|l|c|c|c|c|}
\hline
\textbf{Model Name}  & \textbf{Learning Rate} & \textbf{Momentum} & \textbf{Tradeoff Alignment Loss} & \textbf{Others} \\ \hline
Baseline          & \{\textbf{0.01}, 0.05, 0.25\}   & \{0.5, \textbf{0.9}\}      & -                                & -                                                                                           \\ \hline
AdaBN          & \{{0.01}, 0.05, \textbf{0.25}\}   & \{0.5, \textbf{{0.9}}\}      & -                                & -                                                                                           \\ \hline
MK-MMD               & \{0.01, \textbf{0.05}, 0.25\}   & \{\textbf{0.5}, 0.9\}      & \{0.003, \textbf{0.01}, 0.03\}            & 
$\{2,5,\textbf{10}\}^1$  \\ \hline
DANN                 & \{0.01, \textbf{0.05}, 0.25\}   & \{\textbf{0.5}, 0.9\}      & \{0.1, \textbf{0.3}, 1\}                  & -                                                                                           \\ \hline
OPS-DANN Hard     & \{\textbf{0.05}\}               & \{\textbf{0.5}\}           & \{0.03, 0.1, 0.3, \textbf{1}, 3\}            & -                                                                                           \\ \hline
OPS-DANN Soft      & \{\textbf{0.05}\}               & \{\textbf{0.5}\}           & \{0.1, \textbf{1}, 3\}                    & $\{0.03, \textbf{0.1}, 0.3\}^{2}$                                                                          \\ \hline
Multi-class OPS-DANN & \{\textbf{0.05}\}               & \{\textbf{0.5}\}           & \{0.1, \textbf{0.3}, 1\}                  & -        
\\ \hline
\end{tabular}}
\caption[Grid-Search]{Range of the hyperparameter search for all considered DA-methods.The optimal hyperparameters found on the $S \rightarrow L$ adaptation task are emphasized.}
\label{tab:gridsearch}
\end{table*}

Before training, all models were initialized using Xavier normal initialization \cite{glorot2010understanding}. All the data from each domain are considered during training for each adaptation task, and no test-training split is used.

In this research, several consecutive measurement points are combined into one input sample as input to the model using sequences of length 50 from the multivariate time series with a step size of one.

Model updates were performed using batch gradient descent with batches of size 256. One batch of source and target domain data was processed for each training step.

All DA methods were trained for the same number of epochs for each of the adaptation tasks to ensure a fair comparison. For the $S \rightarrow L$ and $M \rightarrow L$ adaptation tasks, the models were trained for 15 epochs. However, in the $S \rightarrow M$ adaptation task, the models were trained for 25 epochs due to the lower number of model updates per epoch.

The baseline was trained for 40 epochs in case the source domain was short flights  $S$, and for 20 epochs in case the source domain was medium flights $M$.

All DA methods based on DANN additionally required the definition of the reverse gradient factor $\rho$. For this work, the same update rule for $\rho$ is used as the one proposed in the original DANN \cite{ganin2016domain}, which gradually updates $\rho$ from 0 to 1 according to Equation \ref{eq:align_int}. 

\begin{equation}
    \color{black}
    \rho = \frac{2}{1 + e^{-10  \: l_p}} - 1
    \label{eq:align_int}
\end{equation}

The learning rate is reduced after each epoch to ensure a smooth convergence using a learning rate schedule, likewise adopted from the original DANN paper \cite{ganin2016domain}. The proposed schedule is shown in Equation : 

\begin{equation}
    \color{black}
    \alpha = \frac{\alpha_0}{(1 + 10 \: l_p)^{0.75}}
    \label{eq:learningschedule}
\end{equation}

where \textcolor{black}{$l_p$} once again describes the linear training progress. Unlike the gradient reversal factor, which only applies to DANN-based models, the learning rate schedule is used for all models. The initial learning rate $\alpha_{0}$ is found using a hyperparameter search.

The hyperparameters and training specifications mentioned above were primarily selected based on prior work and applied to all models without further refinement. Other hyperparameters were explicitly tuned for each domain adaptation method, such as the learning rate, momentum, and the trade-off between multiple loss objectives. Table~\ref{tab:gridsearch} summarizes the performed grid searches for each method and indicates the optimal hyperparameters found for each method. It is important to emphasize that for all methods, the grid search was only performed on the $S \rightarrow L$ adaptation task. The resulting optimal hyperparameters were then used for the other two adaptation tasks. The $S \rightarrow L$ task was selected to perform the grid search because it has the largest domain gap and is considered as the most challenging task. For AdaBN, the learning rate and momentum found on the $S \rightarrow L$ adaptation task were likely too big for the $M \rightarrow L$ adaptation task and did not lead to convergence. Only in this case, the second best set of hyperparameters found on the $S \rightarrow L$ was used for the $M \rightarrow L$ task, which is also indicated in Table~\ref{tab:gridsearch}.


\subsection{Evaluation Metrics}

In this work, all experiments were evaluated using the two common evaluation metrics used for RUL prediction \cite{saxena2008damage}: the root mean square error (RMSE) and the NASA scoring
function. 

The RMSE is defined as follows:

\begin{equation}
    RMSE = \sqrt{\frac{1}{n_{T}} \sum_{i = 1}^{n_{T}} (\hat{y}^{i} - y^{i})^{2}}
    \label{eq:rmse}
\end{equation}

NASA's scoring metric is not symmetric and penalizes over-estimation more than under-estimation and is defined as follows:

\begin{align}
\begin{split}
    s = \sum_{i = 1}^{n_{T}} \exp{(\alpha \lvert \hat{y}^{i} - y^{i} \lvert)} \\
     \alpha =  \left\{ 
  \begin{array}{ c l }
    1/10 & \quad \textrm{if } \hat{y} - y \geq 0 \\
    1/13              & \quad \textrm{otherwise}
  \end{array}
\right.
\end{split}
\end{align}

\subsection{Representation Transferability}
\label{sec:pad}

The \textit{Proxy A-distance (PAD)} proposed in \cite{ben2010theory}  is a suitable metric to measure the divergence for domain adaptation tasks. Thus, we can use it to evaluate the transferability of representations
Using the PAD, the divergence between two domains is computed by evaluating how well a classifier can separate the source from the target domain. If separation is easy, the two domains are likely dissimilar, and their discrepancy is large. On the other hand, if samples can hardly be discriminated between the two domains, they will likely have a small discrepancy. Using the classification error $\epsilon$, the PAD $\hat{d}_{\mathcal{A}}$ can be found according to Equation \ref{eq:pad}.  

\begin{equation}
    \hat{d}_{\mathcal{A}} = 2 \: (1 - 2\epsilon)
    \label{eq:pad}
\end{equation}

In general, all DA methods that contain a domain-invariant feature learning part aim somewhat to minimize the divergence between the source and the target domain. For example, DANN uses the PAD as a discrepancy measure and aims to minimize it during training in the alignment component \cite{ganin2016domain}.

\section{Results}
\label{sec:Results}

\subsection{Experimental Results}

In this section, we compare the prediction accuracy of the considered models in terms of the RMSE and the NASA score. The results are shown in Table~\ref{tab:results_summarized}. Additionally, the upper bound RMSE for the medium-haul and long-haul flight domains are 3.16 and 2.17, respectively. 

\begin{figure}
    \centering
    \includegraphics[width=1\linewidth]{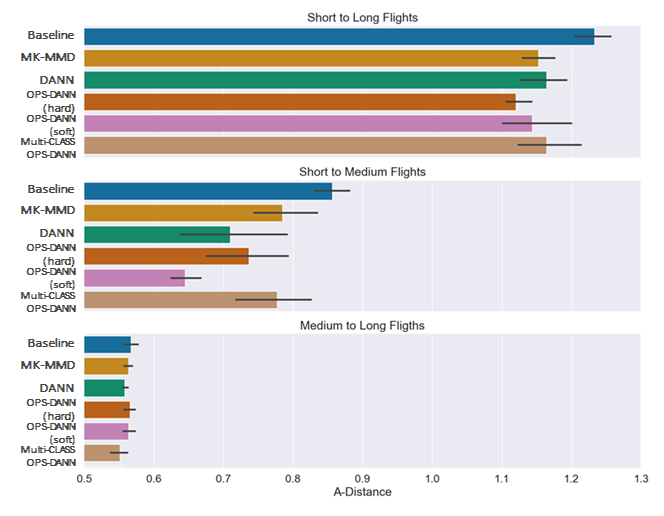}
    \caption[Proxy A-Distance]{A-distance for each method and task respectively}
    \label{fig:a_distance}
\end{figure}

In addition, to evaluating the performance, we also evaluate the PAD (qualitatively measuring the divergence between the domains), calculated from the learned feature embeddings. As described in the section \ref{sec:pad}, the PAD is used to evaluate the ability of the applied DA methods to extract domain-invariant features. In Figure \ref{fig:a_distance}, the A-distance varies from one task to another, showing a larger domain gap for the task $S \rightarrow L$, a smaller one for the task $S \rightarrow M$ and almost no domain gap for the task $M \rightarrow L$

\begin{table*}[ht]
\centering
\caption[Summarized Results]{The results for the three considered Domain Adaptation tasks.}
\small
\scalebox{0.7}{
\begin{tabular}{llllllllllllll}
\cline{1-13}
\multicolumn{1}{|c|}{\multirow{3}{*}{Name}} & \multicolumn{4}{c|}{$\mathbf{S \rightarrow L}$ } & \multicolumn{4}{c|}{$\mathbf{S \rightarrow M}$} & \multicolumn{4}{c|}{$\mathbf{M \rightarrow L}$} & \multicolumn{1}{c}{} \\ \cline{2-13}
\multicolumn{1}{|c|}{} & \multicolumn{2}{c|}{\multirow{2}{*}{\begin{tabular}[c]{@{}c@{}}RMSE \\ Mean $\pm$ 
  Std\end{tabular}}} & \multicolumn{2}{c|}{\multirow{2}{*}{\begin{tabular}[c]{@{}c@{}}s-score   \\ Mean $\pm$   Std\end{tabular}}} & \multicolumn{2}{c|}{\multirow{2}{*}{\begin{tabular}[c]{@{}c@{}}RMSE   \\ Mean $\pm$   Std\end{tabular}}} & \multicolumn{2}{c|}{\multirow{2}{*}{\begin{tabular}[c]{@{}c@{}}s-score  \\ Mean $\pm$   Std\end{tabular}}} & \multicolumn{2}{c|}{\multirow{2}{*}{\begin{tabular}[c]{@{}c@{}}RMSE \\ Mean $\pm$   Std\end{tabular}}} & \multicolumn{2}{c|}{\multirow{2}{*}{\begin{tabular}[c]{@{}c@{}}s-score   \\ Mean $\pm$   Std\end{tabular}}} & \\

\multicolumn{1}{|c|}{} & \multicolumn{2}{c|}{}  & \multicolumn{2}{c|}{}  & \multicolumn{2}{c|}{}  & \multicolumn{2}{c|}{}  & \multicolumn{2}{c|}{} & \multicolumn{2}{c|}{}  & \\ \cline{1-13}

\multicolumn{1}{|l|}{Baseline} & \multicolumn{2}{c|}{7.34 $\pm$ 2.12}   & \multicolumn{2}{c|}{1.74 $\pm$ 0.24}  & \multicolumn{2}{c|}{ 4.59 $\pm$ 0.51}  & \multicolumn{2}{c|}{1.38 $\pm$ 0.07} & \multicolumn{2}{c|}{{2.63 $\pm$ 0.19}} &  \multicolumn{2}{c|}{1.24 $\pm$ 0.03} &  \\ 

\multicolumn{1}{|l|}{AdaBN} & \multicolumn{2}{c|}{7.13 $\pm$ 0.72} & \multicolumn{2}{c|}{1.82 $\pm$ 0.18} & \multicolumn{2}{c|}{5.03 $\pm$ 0.66 } & \multicolumn{2}{c|}{1.49 $\pm$ 0.09}   & \multicolumn{2}{c|}{3.23 $\pm$ 0.26 } & \multicolumn{2}{c|}{1.26 $\pm$ 0.02} & \\ 

\multicolumn{1}{|l|}{MK-MMD} & \multicolumn{2}{c|}{4.76 $\pm$ 0.57} & \multicolumn{2}{c|}{1.47 $\pm$ 0.04} & \multicolumn{2}{c|}{4.40 $\pm$ 0.42 } & \multicolumn{2}{c|}{1.36 $\pm$ 0.04}   & \multicolumn{2}{c|}{2.64 $\pm$ 0.20} & \multicolumn{2}{c|}{1.22 $\pm$ 0.04} & \\ 

\multicolumn{1}{|l|}{DANN} & \multicolumn{2}{c|}{4.38 $\pm$ 0.56} & \multicolumn{2}{c|}{1.43 $\pm$ 0.11} & \multicolumn{2}{c|}{4.07 $\pm$ 0.23} & \multicolumn{2}{c|}{1.35 $\pm$ 0.04}   & \multicolumn{2}{c|}{2.75 $\pm$ 0.35} & \multicolumn{2}{c|}{$\mathbf{1.20 \pm 0.01}$} & \\  \cline{1-13}

\multicolumn{1}{|l|}{\textbf{OPS-DANN Hard}} & \multicolumn{2}{c|}{$\mathbf{3.90 \pm 0.56}$} & \multicolumn{2}{c|}{$\mathbf{1.39 \pm 0.10}$} & \multicolumn{2}{c|}{$\mathbf{3.71 \pm 0.20}$} & \multicolumn{2}{c|}{$\mathbf{1.31 \pm 0.04}$}   & \multicolumn{2}{c|}{2.66 $\pm$ 0.23} & \multicolumn{2}{c|}{$\mathbf{1.20 \pm 0.00}$}& \\  

\multicolumn{1}{|l|}{\textbf{OPS-DANN Soft}} & \multicolumn{2}{c|}{4.00 $\pm$ 0.33} & \multicolumn{2}{c|}{1.40 $\pm$ 0.11} & \multicolumn{2}{c|}{{3.86 $\pm$ 0.18}} & \multicolumn{2}{c|}{1.33 $\pm$ 0.03}   & \multicolumn{2}{c|}{2.72 $\pm$ 0.20} & \multicolumn{2}{c|}{1.21 $\pm$ 0.02} & \\  

\multicolumn{1}{|l|}{\begin{tabular}[c]{@{}l@{}}\textbf{Multi-Class} \\ \textbf{OPS-DANN} \end{tabular}} & \multicolumn{2}{c|}{{3.97 $\pm$ 0.65}} & \multicolumn{2}{c|}{1.40 $\pm$ 0.11} & \multicolumn{2}{c|}{3.93 $\pm$ 0.46} & \multicolumn{2}{c|}{1.33 $\pm$ 0.03}   & \multicolumn{2}{c|}{$\mathbf{2.58 \pm 0.20}$} & \multicolumn{2}{c|}{$\mathbf{1.20 \pm 0.01}$} & \\  \cline{1-13}

\end{tabular}}
\label{tab:results_summarized}
\end{table*}

\subsubsection{\texorpdfstring{${S \rightarrow L}$ task}{Lg}} 

 \noindent\textbf{Performance evaluation:} The adaptation task from the short to the long flight domain is the most challenging because it has the largest domain gap. The baseline model trained solely on source data for this task (without any adaptation)  exhibits an RMSE value that is  240$\%$ larger than the one of the models trained on the target domain. As seen in Table \ref{tab:results_summarized}, the MK-MMD and DANN methods improved the baseline methods in terms of RMSE by 35 and 40 $\%$, respectively. The improvements by DA methods are also visible in the NASA score. AdaBN is a simple method without any additional parameters. It improves the model performance with respect to RMSE only slightly. However, it leads to a higher NASA score compared to the baseline due to the over-estimations of the RUL predictions. For the $S \rightarrow L$ task, all three proposed OPS-DANN variants improve substantially upon the baseline by at least $45\%$ in terms of RMSE and by $20\%$ in terms of S-score. 
 
 It is similarly important to compare the performance of a novel DA method to other state-of-the-art methods. Even though DANN has the lowest RMSE out of the three considered baseline methods, the proposed OPS-DANN methods outperform DANN by at least $9\%$ in RMSE and $2\%$ in S-score.
Thus, all models benefit from unlabeled target domain data, as expected in this setup. However, comparisons with the upper bound RMSE of 2.17 show that by using labeled short-haul flight data for adaptation, there is still a considerable performance gap to models trained with the labeled long-haul flight data. 

  \noindent\textbf{Evaluation of the alignment:} Figure \ref{fig:a_distance}, shows that all three proposed OPS-DANN methods can extract features that contain less discriminative information about the two domains than the baseline on the $S \rightarrow M$ DA task. The two methods using individual domain discriminators (OPS-DANN soft and hard) reach the lowest PAD values for this task, indicating that the usage of a dedicated domain discriminator per operating phase can support the extraction of features that are more domain-invariant compared to domain adaptation techniques aligning the entire marginal distributions of the two domains.

  \begin{figure*}[h]
\centering
  \begin{subfigure}[b]{.23\linewidth}
    \centering
    \includegraphics[width=.99\textwidth]{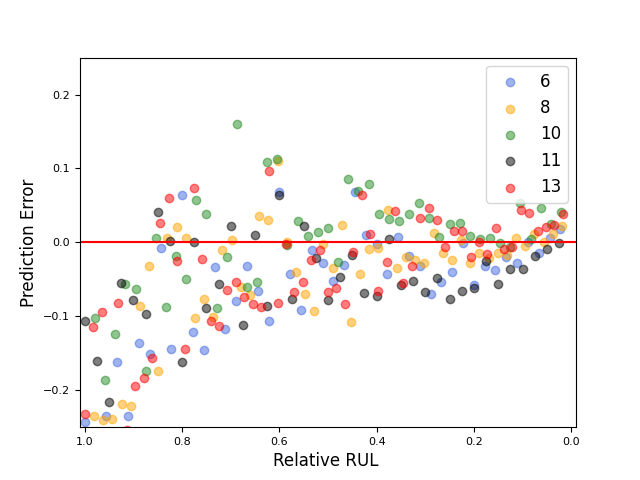}
    \caption{$S \rightarrow L$ DANN}\label{fig:1a}
  \end{subfigure}%
  \begin{subfigure}[b]{.23\linewidth}
    \centering
    \includegraphics[width=.99\textwidth]{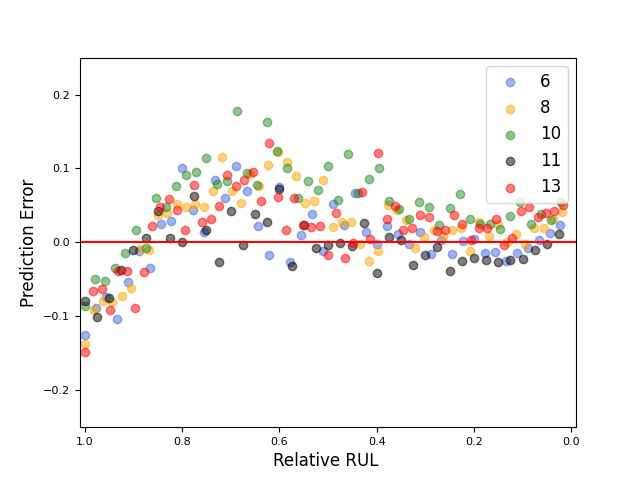}
    \caption{$S \rightarrow L$ OPS-DANN Hard}\label{fig:1b}
  \end{subfigure}%
  \begin{subfigure}[b]{.23\linewidth}
    \centering
    \includegraphics[width=.99\textwidth]{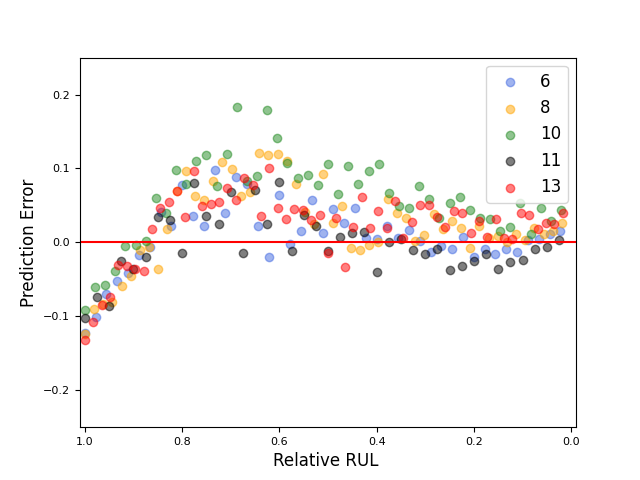}
    \caption{$S \rightarrow L$ OPS-DANN Soft}\label{fig:1c}
  \end{subfigure}%
  \begin{subfigure}[b]{.23\linewidth}
    \centering
    \includegraphics[width=.99\textwidth]{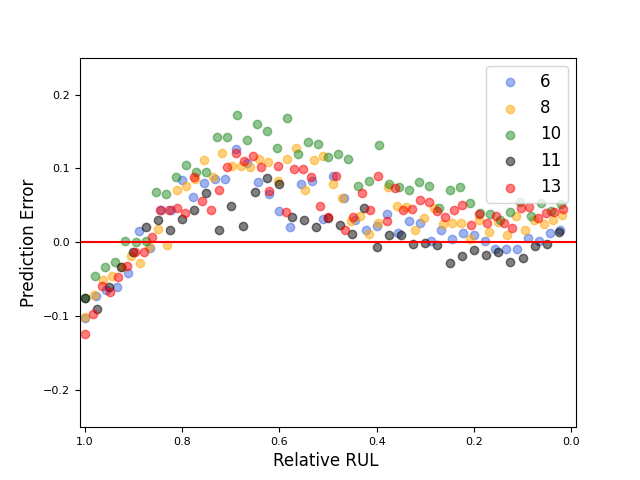}
    \caption{$S \rightarrow L$ Multi-Class OPS}\label{fig:1d}
  \end{subfigure}\\%
  \begin{subfigure}[b]{.23\linewidth}
    \centering
    \includegraphics[width=.99\textwidth]{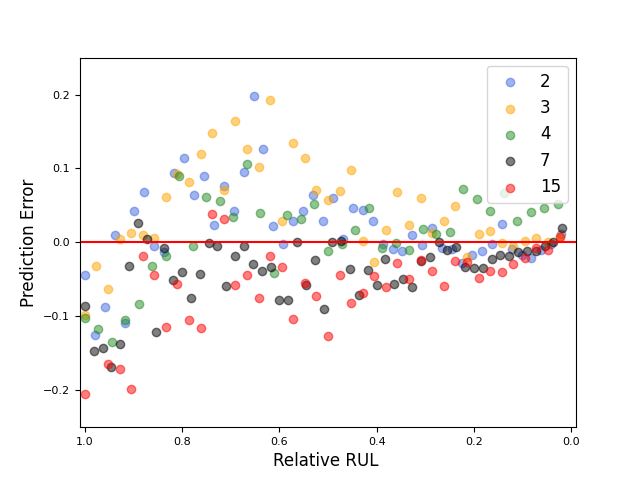}
    \caption{$S \rightarrow M$ DANN}\label{fig:1e}
  \end{subfigure}%
  \begin{subfigure}[b]{.23\linewidth}
    \centering
    \includegraphics[width=.99\textwidth]{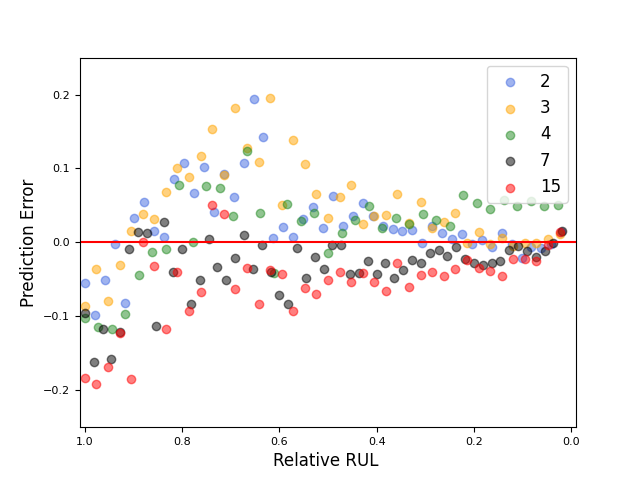}
    \caption{$S \rightarrow M$ OPS-DANN Hard}\label{fig:1f}
  \end{subfigure}%
  \begin{subfigure}[b]{.23\linewidth}
    \centering
    \includegraphics[width=.99\textwidth]{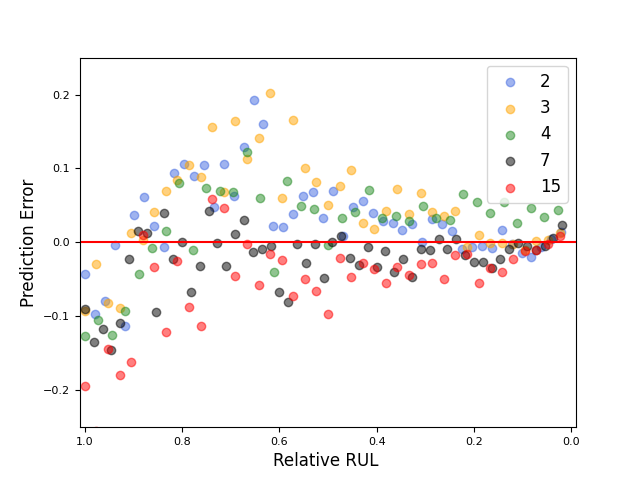}
    \caption{$S \rightarrow M$ OPS-DANN Soft}\label{fig:1g}
  \end{subfigure}%
  \begin{subfigure}[b]{.23\linewidth}
    \centering
    \includegraphics[width=.99\textwidth]{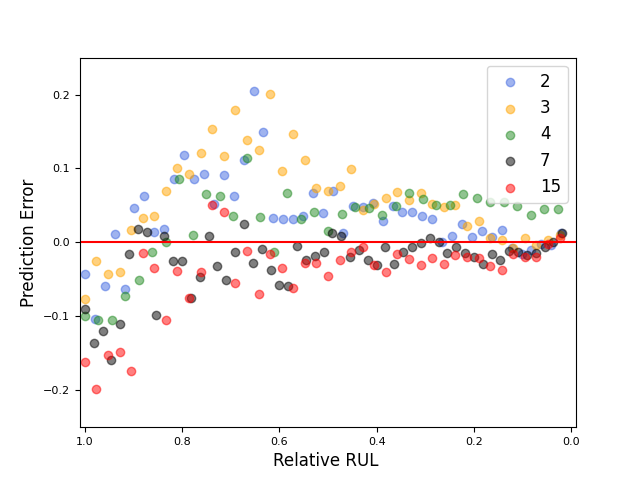}
    \caption{$S \rightarrow M$ Multi-Class OPS}\label{fig:1h}
  \end{subfigure}\\%
  \begin{subfigure}[b]{.23\linewidth}
    \centering
    \includegraphics[width=.99\textwidth]{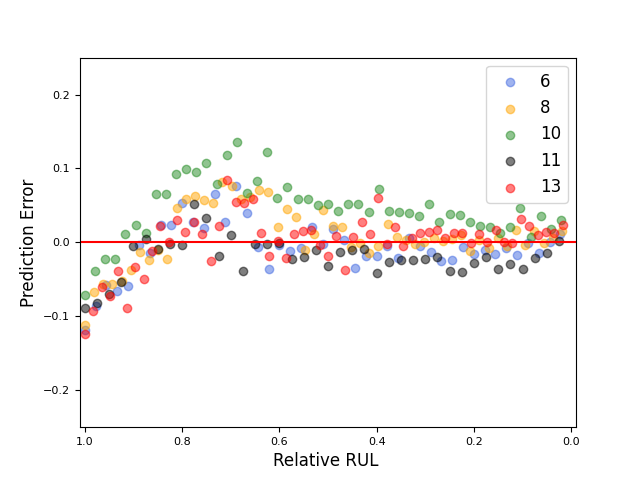}
    \caption{$M \rightarrow L$ DANN}\label{fig:1i}
  \end{subfigure}%
  \begin{subfigure}[b]{.23\linewidth}
    \centering
    \includegraphics[width=.99\textwidth]{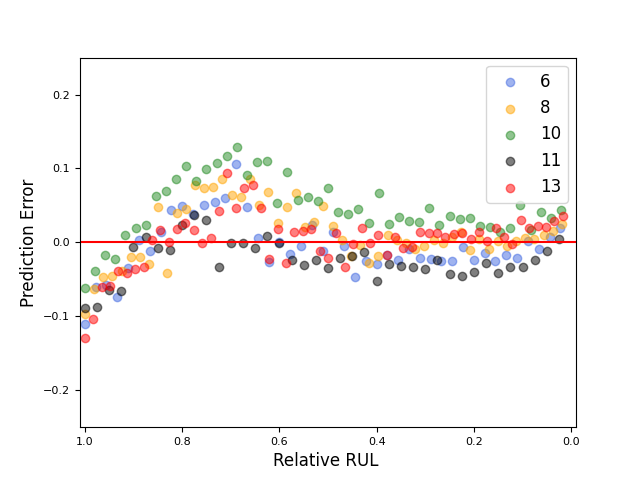}
    \caption{$M \rightarrow L$ OPS-DANN Hard}\label{fig:1j}
  \end{subfigure}%
  \begin{subfigure}[b]{.23\linewidth}
    \centering
    \includegraphics[width=.99\textwidth]{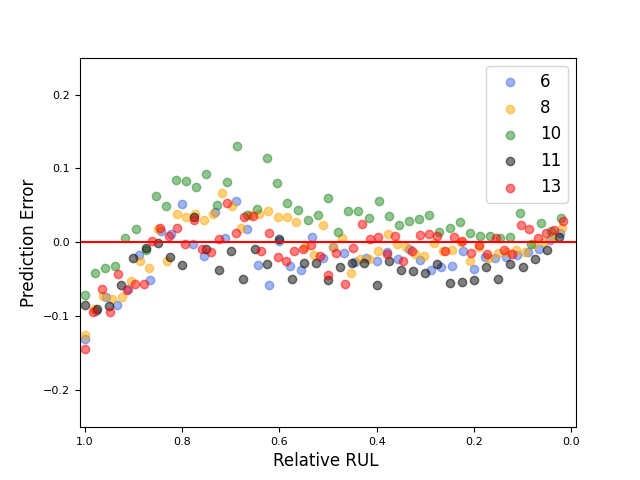}
    \caption{$M \rightarrow L$ OPS-DANN Soft}\label{fig:1k}
  \end{subfigure}%
  \begin{subfigure}[b]{.23\linewidth}
    \centering
    \includegraphics[width=.99\textwidth]{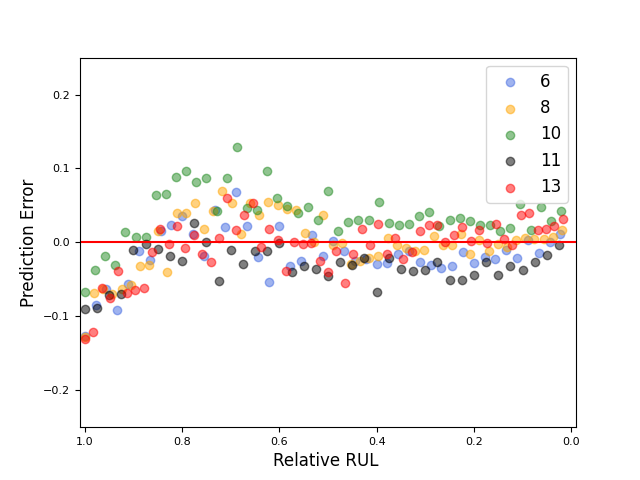}
    \caption{$M \rightarrow L$ Multi-Class OPS}\label{fig:1l}
  \end{subfigure}\\%
  \caption{Prediction error of the different models as a function of the
relative lifetime.}\label{fig:1}
\end{figure*}
  
\subsubsection{\texorpdfstring{${S \rightarrow M}$ task}{Lg}} 

  \noindent\textbf{Performance evaluation:} Compared to the $S \rightarrow L$ task, the $S \rightarrow M$ task has a smaller domain gap. The baseline model trained without any adaptation solely based on source data for this task results in an RMSE that is 45$\%$ larger than the one from the model trained on the target domain. Due to the smaller domain gap in the $S \rightarrow M$ task compared to the $S \rightarrow L$ task, the source domain is more similar to the target domain and, therefore, it allows learning models with a similar performance level as models trained on the target data. As seen in Table \ref{tab:results_summarized}, the MK-MMD and DANN methods are able to improve the baseline methods slightly in terms of RMSE by 4 and 11$\%$, respectively. However, in terms of the s-score, the results decreased slightly from 1.38 to 1.36 and 1.35, respectively. AdaBN approach performed worse than the baseline in this case. Also, in this task, DANN is the best-performing comparison model. Nevertheless, all three proposed OPS-DANN methods outperform the original DANN. While for the Multi-Class OPS-DANN, the improvement is 3\% compared to the DANN performance, the OPS-DANN (soft) and OPS-DANN (hard) reduce the RMSE by $5\%$ and $9\%$, respectively, and the s-score  by $1\%$  and  $3\%$
  
  \noindent\textbf{Evaluation of the alignment:} On the $S \rightarrow M$ task, all methods reach considerably lower PAD values, even the lower bound, without any adaptation. This makes it more challenging to differentiate between these two domains. However, it is worth noting that all the proposed methods in the Operation Profile-specific Domain Adaptation Network (OPS-DANN) have considerably lower PAD values compared to the baseline. However, only the OPS-DANN (soft) method can extract more domain-invariant features than DANN, while the other two proposed methods have slightly higher PAD values. Even though OPS-DANN (hard) has a higher PAD than both DANN and OPS-DANN (soft), it reaches a lower RMSE demonstrating that the sole ability to extract domain-invariant features does not guarantee a superior performance on the regression task.

\subsubsection{\texorpdfstring{${M \rightarrow L}$ task}{Lg}} 

  \noindent\textbf{Performance evaluation:} The third considered task uses medium-haul flights as the source domain and aims to adapt a model to perform well on long-haul flights. Compared to the two previous tasks, the discrepancy between these two domains is much smaller. This observation is also confirmed by the performance of the baseline method that has a value RMSE that is only 21$\%$ larger than the one of a model trained on the target domain. All domain adaptation methods performed slightly worse than the baseline model. Using the MK-MMD and DANN methods appears to result in under-estimated predictions, thus achieving a lower s-score than the baseline.
Out of the three proposed OPS-DANN models, only one of them (Multi-Class OPS-DANN) has a slightly lower RMSE value compared to the baseline (2.58 compared to 2.63). The other two approaches are at a similar performance level as the baseline model, however, achieving a lower s-score (1.20 compared to 1.24). The observed slight differences between the models are likely insignificant and are potentially caused by the randomness of the training procedure. Overall, none of the comparison methods DA methods could improve this task's target-free baseline significantly. None of the methods are able to reach the upper bound RMSE value of 2.17, which is achieved when the model is trained on the labeled target dataset.

\textcolor{black}{
\noindent\textbf{Visualizations:} We provide additional visualizations, in Figure \ref{fig:1}, by showing the test  RUL evolution over time for the three adaptation approaches ($S \rightarrow L$, $S \rightarrow M$, $M \rightarrow L$) using DANN and the proposed three methods. The figures confirm our initial expectations. As the units approach the end of their lifetime, the predictions tend to converge with the ground truth RUL value.}

  \noindent\textbf{Evaluation of the alignment:}
Comparing the PAD values of all applied methods in Figure \ref{fig:a_distance}, it becomes apparent that none of the methods are able to reduce substantially the domain distinctiveness compared to the models trained without adaptation.

\begin{figure}[ht]
\centering
\begin{subfigure}[t]{0.45\columnwidth}
    \includegraphics[width=\textwidth]{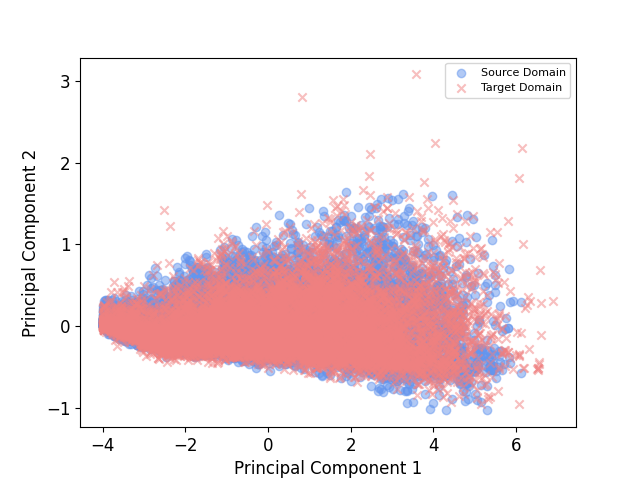}
    \caption{DANN}
\end{subfigure}
\hspace{0mm}
\begin{subfigure}[t]{0.45\columnwidth}
    \includegraphics[width=1\textwidth]{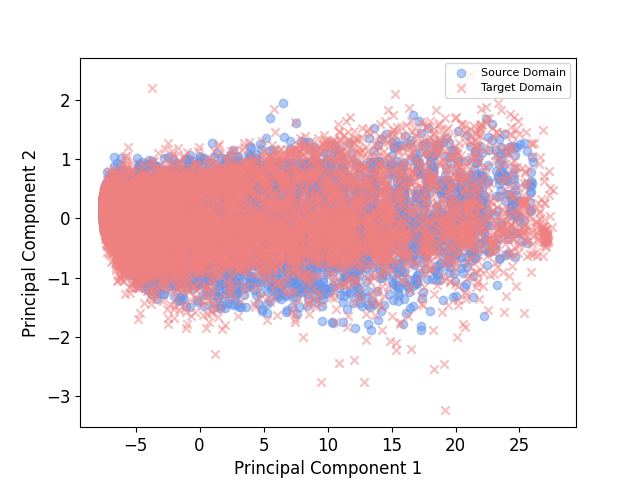}
    \caption{OPS-DANN (hard)} 
\end{subfigure} 
\begin{subfigure}[t]{0.45\columnwidth}
    \includegraphics[width=\textwidth]{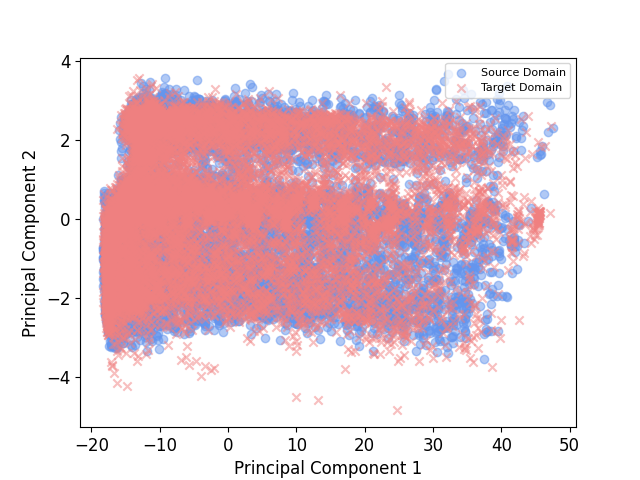}
    \caption{OPS-DANN (soft)} 
\end{subfigure} 
\hspace{0mm}
\begin{subfigure}[t]{0.45\columnwidth}
    \includegraphics[width=1\textwidth]{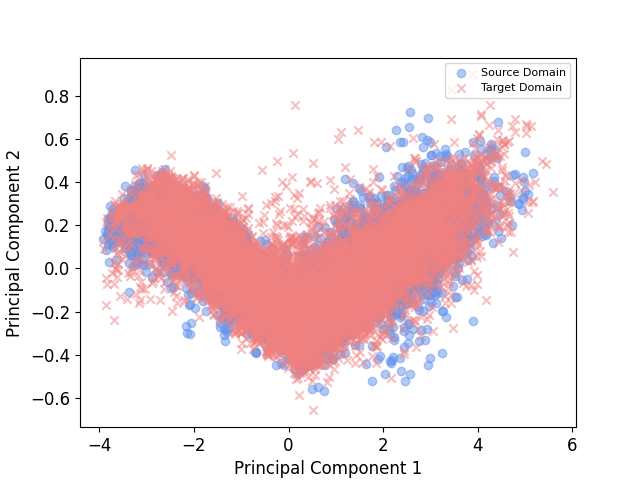}
    \caption{Multi-class \\ OPS-DANN} 
\end{subfigure} 
    \caption[Feature Embeddings Domains]{Visualizing the Impact of the Operation Profile Alignment on \textbf{Domain}  alignment: A Comparison of DANN, OPS-DANN (soft), OPS-DANN (soft), and Multi-Class OPS-DANN Models using PCA on the Embedding space}
    \label{fig:pac_embeddings}
\end{figure}

\begin{figure}[ht]
\centering
\begin{subfigure}[t]{0.45\columnwidth}
    \includegraphics[width=\textwidth]{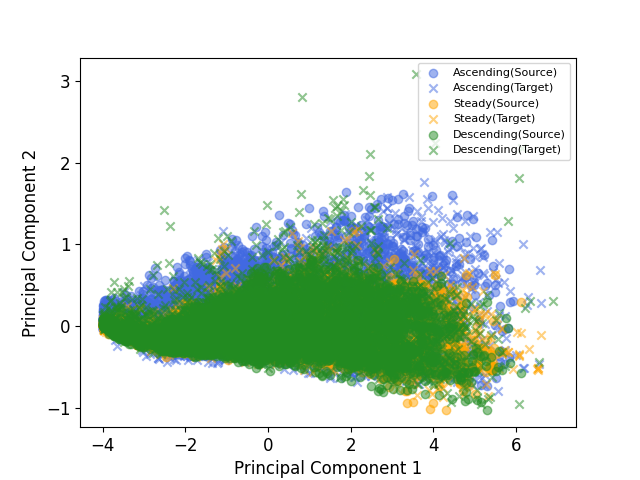}
    \caption{DANN}
\end{subfigure}
\hspace{0mm}
\begin{subfigure}[t]{0.45\columnwidth}
    \includegraphics[width=1\textwidth]{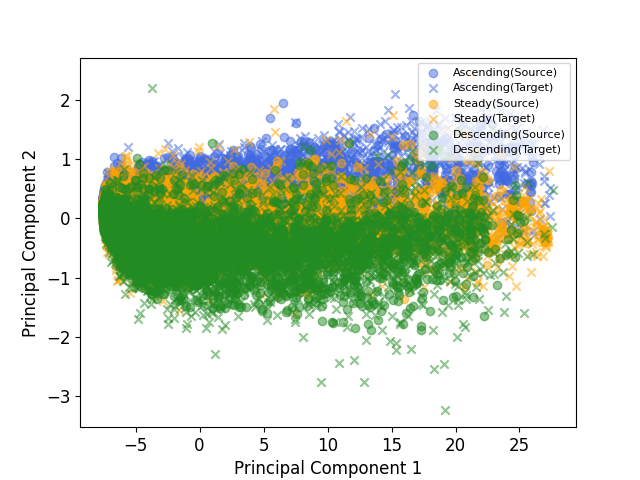}
    \caption{OPS-DANN (hard)} 
\end{subfigure} 
\begin{subfigure}[t]{0.45\columnwidth}
    \includegraphics[width=\textwidth]{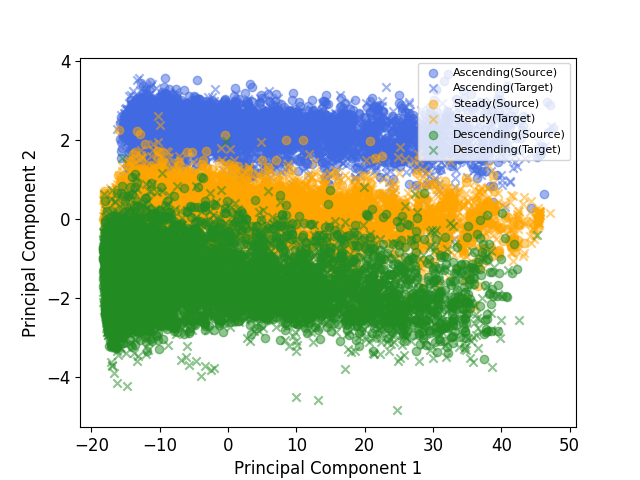}
    \caption{OPS-DANN (soft)} 
\end{subfigure} 
\hspace{0mm}
\begin{subfigure}[t]{0.45\columnwidth}
    \includegraphics[width=1\textwidth]{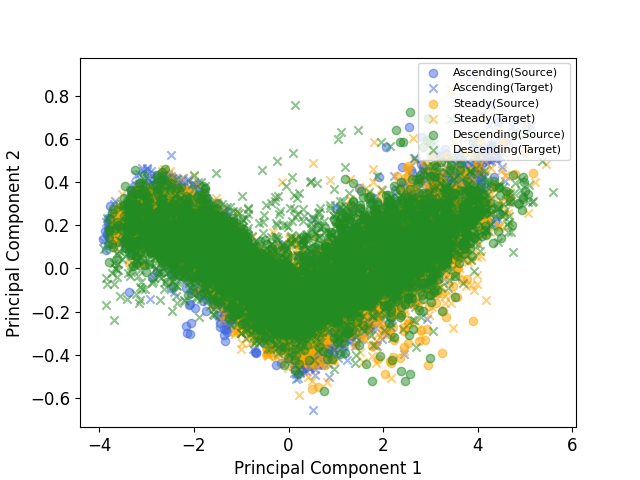}
    \caption{Multi-class \\ OPS-DANN} 
\end{subfigure} 
    \caption[Feature Embeddings]{Visualizing the Impact of the Operation Profile Alignment on the \textbf{Phase} alignment: A Comparison of DANN, OPS-DANN (soft), OPS-DANN (soft), and Multi-Class OPS-DANN Models using PCA on the Embedding space}
    \label{fig:pac_embeddings_op}
\end{figure}

\section{Discussion}


The findings of this study demonstrate that, on two out of the three tasks, the proposed three variants of the operation phase alignment methods perform similarly well and are able to achieve better results compared to both the DANN model as well as the other domain adaptation methods. This demonstrates the effectiveness of the proposed methods in adapting to different domains, particularly in addressing the unique challenges posed by multimodal feature distributions and the variability of operating conditions and system configurations. Furthermore, it should be noted that DANN is already an effective method for domain adaptation. However, the proposed operation phase alignment methods are able to improve upon the performance of DANN and other traditional domain adaptation methods in the context of PHM.

In addition to evaluating the performance of the proposed methods for RUL prediction, we also investigate the impact of each method on the embeddings generated by the alignment process. To enable 2D visualizations, we applied a principal component analysis (PCA) and present the first two principal components in Figures \ref{fig:pac_embeddings} and \ref{fig:pac_embeddings_op}. This visualization provides insights into how the different phases of the operation profile are represented in the embeddings and how the proposed methods affect this representation. 

As shown in Figure \ref{fig:pac_embeddings} DANN and  our proposed approaches are able to effectively align the source and target embeddings, as demonstrated by the overlapping of the source and target data points in the figures. However, it is worth noting that there is a distinct difference between the methods with respect to how distinguishable the operation phases are in the embedding space. 

In Figures \ref{fig:pac_embeddings} and \ref{fig:pac_embeddings_op}, the OPS-DANN soft and hard approaches have phases that can be distinguished but that are more pronounced for the soft approach. On the contrary, the Multi-Class OPS-DANN approach has only one cluster that is similar to the one obtained with the DANN approach which is not able to distinguish between the operating phases. However, the difference between DANN and Multi-Class OPS-DANN is that they have different shapes in the embedding space. Moreover, DANN aims to learn domain invariant features, and Multi-Class OPS-DANN aims to learn domain and operating-condition invariant features. This difference in the ability to distinguish between the operation phases highlights the unique characteristics of each proposed approach and their ability to align the operating characteristics and phases of the different domains.

It should be noted that the Multi-Class OPS-DANN uses the operation phase labels in a distinctly different way than the other two proposed methods. Instead of aligning the marginal distributions of all three operation phases separately, it aims to extract features invariant to both the domain and the operation phase of the input data simultaneously. Figure \ref{fig:pac_embeddings_op} shows that Multi-Class OCS-DANN successfully learns such a feature representation that is invariant to the operation phases.

Contrary to the Multi-Class OPS-DANN, the OCS-DANN (soft) models contain an additional classifier that aims to learn how to distinguish between different operating phases while simultaneously learning to extract features that are domain invariant. Figure \ref{fig:pac_embeddings_op} illustrates this behavior: each operating phase forms a separate cluster while the source and target domains overlap to a large extent. This behavior is similar to the OCS-DANN (hard).

This observation highlights the effectiveness of the proposed approach in aligning the operating characteristics and conditions of the different domains, leading to improved performance for RUL prediction

\section{Conclusion}
\label{sec:Conclusion}


In this paper, we propose a novel approach that utilizes domain adaptation techniques to align the operating phases to improve the accuracy of RUL predictions. 

The main novelty of our proposed approach is the integration of the information on the different phases of the operation profile into the alignment process. The proposed approaches align the marginal distributions of each phase of the operation profile in the labeled source domain with its counterpart in the unlabeled target domain. Two novel domain adaptation approaches are proposed based on an adversarial domain adaptation by considering the different phases of the operation profile separately.

The proposed methods have shown to be effective in improving the performance of deep learning models for RUL prediction by effectively transferring the models between sub-fleets that are operated under different conditions. The results of this study demonstrate the potential of these methods to improve the accuracy and reliability of prognostics and health management in real-world applications. Furthermore, the proposed methods have a better performance compared to state-of-the-art domain adaptation methods such as DANN, MK-MMD, and AdaBN. 

There are three interesting potential future research directions resulting from this research. First, the proposed methods can be extended to enable them to tackle challenges arising from imbalanced operating conditions, which is a very common case in practical applications. Second, it would be interesting to investigate whether learning a soft assignment before performing adaptation may be beneficial by training an operating phase classifier separately from the DA model. 
Thirdly, while this research focused on regression, the proposed methodology can be easily extended to classification tasks such as for example for fault diagnostics problems. We leave the aforementioned open research directions for future work.

\section{Acknowledgments}
\label{sec:Acknowledgments}

This work was supported by
the Swiss National Science Foundation under Grant PP00P2$\_$176878


\bibliographystyle{elsarticle-num} 
\bibliography{cas-refs}

\begin{thebibliography}{10}
\expandafter\ifx\csname url\endcsname\relax
  \def\url#1{\texttt{#1}}\fi
\expandafter\ifx\csname urlprefix\endcsname\relax\def\urlprefix{URL }\fi
\expandafter\ifx\csname href\endcsname\relax
  \def\href#1#2{#2} \def\path#1{#1}\fi

\bibitem{si2011remaining}
X.-S. Si, W.~Wang, C.-H. Hu, D.-H. Zhou, Remaining useful life estimation--a
  review on the statistical data driven approaches, European journal of
  operational research 213~(1) (2011) 1--14.

\bibitem{zhao2019deep}
R.~Zhao, R.~Yan, Z.~Chen, K.~Mao, P.~Wang, R.~X. Gao, Deep learning and its
  applications to machine health monitoring, Mechanical Systems and Signal
  Processing 115 (2019) 213--237.

\bibitem{fink2020potential}
O.~Fink, Q.~Wang, M.~Svensen, P.~Dersin, W.-J. Lee, M.~Ducoffe, Potential,
  challenges and future directions for deep learning in prognostics and health
  management applications, Engineering Applications of Artificial Intelligence
  92 (2020) 103678.

\bibitem{rombach2021contrastive}
K.~Rombach, G.~Michau, O.~Fink, Contrastive learning for fault detection and
  diagnostics in the context of changing operating conditions and novel fault
  types, Sensors 21~(10) (2021) 3550.

\bibitem{rombach2023controlled}
K.~Rombach, G.~Michau, O.~Fink, Controlled generation of unseen faults for
  partial and open-partial domain adaptation, Reliability Engineering \& System
  Safety 230 (2023) 108857.

\bibitem{wilson2020survey}
G.~Wilson, D.~J. Cook, A survey of unsupervised deep domain adaptation, ACM
  Transactions on Intelligent Systems and Technology (TIST) 11~(5) (2020)
  1--46.

\bibitem{li2020systematic}
C.~Li, S.~Zhang, Y.~Qin, E.~Estupinan, A systematic review of deep transfer
  learning for machinery fault diagnosis, Neurocomputing 407 (2020) 121--135.

\bibitem{wang2019domain}
Q.~Wang, G.~Michau, O.~Fink, Domain adaptive transfer learning for fault
  diagnosis, in: 2019 Prognostics and System Health Management Conference
  (PHM-Paris), IEEE, 2019, pp. 279--285.

\bibitem{da2020remaining}
P.~R. d.~O. da~Costa, A.~Ak{\c{c}}ay, Y.~Zhang, U.~Kaymak, Remaining useful
  lifetime prediction via deep domain adaptation, Reliability Engineering \&
  System Safety 195 (2020) 106682.

\bibitem{cheng2021transferable}
H.~Cheng, X.~Kong, G.~Chen, Q.~Wang, R.~Wang, Transferable convolutional neural
  network based remaining useful life prediction of bearing under multiple
  failure behaviors, Measurement 168 (2021) 108286.

\bibitem{zhuang2021temporal}
J.~Zhuang, M.~Jia, Y.~Ding, P.~Ding, Temporal convolution-based transferable
  cross-domain adaptation approach for remaining useful life estimation under
  variable failure behaviors, Reliability Engineering \& System Safety 216
  (2021) 107946.

\bibitem{ganin2015unsupervised}
Y.~Ganin, V.~Lempitsky, Unsupervised domain adaptation by backpropagation, in:
  International conference on machine learning, PMLR, 2015, pp. 1180--1189.

\bibitem{arjovsky2017towards}
M.~Arjovsky, L.~Bottou, Towards principled methods for training generative
  adversarial networks, arXiv preprint arXiv:1701.04862 (2017).

\bibitem{long2018conditional}
M.~Long, Z.~Cao, J.~Wang, M.~I. Jordan, Conditional adversarial domain
  adaptation, Advances in neural information processing systems 31 (2018).

\bibitem{ganin2016domain}
Y.~Ganin, E.~Ustinova, H.~Ajakan, P.~Germain, H.~Larochelle, F.~Laviolette,
  M.~Marchand, V.~Lempitsky, Domain-adversarial training of neural networks,
  The journal of machine learning research 17~(1) (2016) 2096--2030.

\bibitem{zamir2018taskonomy}
A.~R. Zamir, A.~Sax, W.~Shen, L.~J. Guibas, J.~Malik, S.~Savarese, Taskonomy:
  Disentangling task transfer learning, in: Proceedings of the IEEE conference
  on computer vision and pattern recognition, 2018, pp. 3712--3722.

\bibitem{Wang_2021_ICCV}
Q.~Wang, D.~Dai, L.~Hoyer, L.~Van~Gool, O.~Fink, Domain adaptive semantic
  segmentation with self-supervised depth estimation, in: Proceedings of the
  IEEE/CVF International Conference on Computer Vision (ICCV), 2021, pp.
  8515--8525.

\bibitem{arias2021aircraft}
M.~Arias~Chao, C.~Kulkarni, K.~Goebel, O.~Fink, Aircraft engine run-to-failure
  dataset under real flight conditions for prognostics and diagnostics, Data
  6~(1) (2021) 5.

\bibitem{pan2009survey}
S.~J. Pan, Q.~Yang, A survey on transfer learning, IEEE Transactions on
  knowledge and data engineering 22~(10) (2009) 1345--1359.

\bibitem{long2015learning}
M.~Long, Y.~Cao, J.~Wang, M.~Jordan, Learning transferable features with deep
  adaptation networks, in: International conference on machine learning, PMLR,
  2015, pp. 97--105.

\bibitem{long2017deep}
M.~Long, H.~Zhu, J.~Wang, M.~I. Jordan, Deep transfer learning with joint
  adaptation networks (2017).
\newblock \href {http://arxiv.org/abs/1605.06636} {\path{arXiv:1605.06636}}.

\bibitem{saito2018maximum}
K.~Saito, K.~Watanabe, Y.~Ushiku, T.~Harada, Maximum classifier discrepancy for
  unsupervised domain adaptation, in: Proceedings of the IEEE conference on
  computer vision and pattern recognition, 2018, pp. 3723--3732.

\bibitem{li2016revisiting}
Y.~Li, N.~Wang, J.~Shi, J.~Liu, X.~Hou, Revisiting batch normalization for
  practical domain adaptation, arXiv preprint arXiv:1603.04779 (2016).

\bibitem{carlucci2017autodial}
F.~M. Carlucci, L.~Porzi, B.~Caputo, E.~Ricci, S.~R. Bulo, Autodial: Automatic
  domain alignment layers, in: 2017 IEEE international conference on computer
  vision (ICCV), IEEE, 2017, pp. 5077--5085.

\bibitem{cortes2011domain}
C.~Cortes, M.~Mohri, Domain adaptation in regression, in: International
  Conference on Algorithmic Learning Theory, Springer, 2011, pp. 308--323.

\bibitem{mansour2009domain}
Y.~Mansour, M.~Mohri, A.~Rostamizadeh, Domain adaptation: Learning bounds and
  algorithms (2009).
\newblock \href {http://arxiv.org/abs/0902.3430} {\path{arXiv:0902.3430}}.

\bibitem{chen2021representation}
X.~Chen, S.~Wang, J.~Wang, M.~Long, Representation subspace distance for domain
  adaptation regression, in: International Conference on Machine Learning,
  PMLR, 2021, pp. 1749--1759.

\bibitem{nejjar2023dare}
I.~Nejjar, Q.~Wang, O.~Fink, Dare-gram: Unsupervised domain adaptation
  regression by aligning inverse gram matrices, in: Proceedings of the IEEE/CVF
  Conference on Computer Vision and Pattern Recognition, 2023, pp.
  11744--11754.

\bibitem{li2022perspective}
W.~Li, R.~Huang, J.~Li, Y.~Liao, Z.~Chen, G.~He, R.~Yan, K.~Gryllias, A
  perspective survey on deep transfer learning for fault diagnosis in
  industrial scenarios: Theories, applications and challenges, Mechanical
  Systems and Signal Processing 167 (2022) 108487.

\bibitem{zhang2020unsupervised}
Z.~Zhang, H.~Chen, S.~Li, Z.~An, Unsupervised domain adaptation via enhanced
  transfer joint matching for bearing fault diagnosis, Measurement 165 (2020)
  108071.

\bibitem{pan2019approach}
Y.~Pan, F.~Mei, H.~Miao, J.~Zheng, K.~Zhu, H.~Sha, An approach for hvcb
  mechanical fault diagnosis based on a deep belief network and a transfer
  learning strategy, Journal of Electrical Engineering \& Technology 14~(1)
  (2019) 407--419.

\bibitem{michau2021unsupervised}
G.~Michau, O.~Fink, Unsupervised transfer learning for anomaly detection:
  Application to complementary operating condition transfer, Knowledge-Based
  Systems 216 (2021) 106816.

\bibitem{GAO2021356}
Y.~Gao, X.~Liu, H.~Huang, J.~Xiang, A hybrid of fem simulations and generative
  adversarial networks to classify faults in rotor-bearing systems, ISA
  Transactions 108 (2021) 356--366.
\newblock \href {https://doi.org/https://doi.org/10.1016/j.isatra.2020.08.012}
  {\path{doi:https://doi.org/10.1016/j.isatra.2020.08.012}}.

\bibitem{doi:10.1177/1475921720980718}
K.~Yu, Q.~Fu, H.~Ma, T.~R. Lin, X.~Li, Simulation data driven weakly supervised
  adversarial domain adaptation approach for intelligent cross-machine fault
  diagnosis, Structural Health Monitoring 20~(4) (2021) 2182--2198.
\newblock \href {http://arxiv.org/abs/https://doi.org/10.1177/1475921720980718}
  {\path{arXiv:https://doi.org/10.1177/1475921720980718}}, \href
  {https://doi.org/10.1177/1475921720980718}
  {\path{doi:10.1177/1475921720980718}}.

\bibitem{YANG2019692}
B.~Yang, Y.~Lei, F.~Jia, S.~Xing, An intelligent fault diagnosis approach based
  on transfer learning from laboratory bearings to locomotive bearings,
  Mechanical Systems and Signal Processing 122 (2019) 692--706.
\newblock \href {https://doi.org/https://doi.org/10.1016/j.ymssp.2018.12.051}
  {\path{doi:https://doi.org/10.1016/j.ymssp.2018.12.051}}.

\bibitem{wang2021integrating}
Q.~Wang, C.~Taal, O.~Fink, Integrating expert knowledge with domain adaptation
  for unsupervised fault diagnosis, IEEE Transactions on Instrumentation and
  Measurement 71 (2021) 1--12.

\bibitem{lu2016deep}
W.~Lu, B.~Liang, Y.~Cheng, D.~Meng, J.~Yang, T.~Zhang, Deep model based domain
  adaptation for fault diagnosis, IEEE Transactions on Industrial Electronics
  64~(3) (2016) 2296--2305.

\bibitem{li2019multi}
X.~Li, W.~Zhang, Q.~Ding, J.-Q. Sun, Multi-layer domain adaptation method for
  rolling bearing fault diagnosis, Signal processing 157 (2019) 180--197.

\bibitem{zhang2018adversarial}
B.~Zhang, W.~Li, J.~Hao, X.-L. Li, M.~Zhang, Adversarial adaptive 1-d
  convolutional neural networks for bearing fault diagnosis under varying
  working condition, arXiv preprint arXiv:1805.00778 (2018).

\bibitem{bao2021enhanced}
H.~Bao, Z.~Yan, S.~Ji, J.~Wang, S.~Jia, G.~Zhang, B.~Han, An enhanced sparse
  filtering method for transfer fault diagnosis using maximum classifier
  discrepancy, Measurement Science and Technology 32~(8) (2021) 085105.

\bibitem{8543590}
Z.~Tong, W.~Li, B.~Zhang, F.~Jiang, G.~Zhou, Bearing fault diagnosis under
  variable working conditions based on domain adaptation using feature transfer
  learning, IEEE Access 6 (2018) 76187--76197.
\newblock \href {https://doi.org/10.1109/ACCESS.2018.2883078}
  {\path{doi:10.1109/ACCESS.2018.2883078}}.

\bibitem{han2021hybrid}
T.~Han, Y.-F. Li, M.~Qian, A hybrid generalization network for intelligent
  fault diagnosis of rotating machinery under unseen working conditions, IEEE
  Transactions on Instrumentation and Measurement 70 (2021) 1--11.

\bibitem{9136845}
Y.~Li, Y.~Song, L.~Jia, S.~Gao, Q.~Li, M.~Qiu, Intelligent fault diagnosis by
  fusing domain adversarial training and maximum mean discrepancy via ensemble
  learning, IEEE Transactions on Industrial Informatics 17~(4) (2021)
  2833--2841.
\newblock \href {https://doi.org/10.1109/TII.2020.3008010}
  {\path{doi:10.1109/TII.2020.3008010}}.

\bibitem{2020MSSP14506962J}
J.~{Jiao}, M.~{Zhao}, J.~{Lin}, K.~{Liang}, {Residual joint adaptation
  adversarial network for intelligent transfer fault diagnosis}, Mechanical
  Systems and Signal Processing 145 (2020) 106962.
\newblock \href {https://doi.org/10.1016/j.ymssp.2020.106962}
  {\path{doi:10.1016/j.ymssp.2020.106962}}.

\bibitem{aerospace9090516}
W.~Li, W.~Yang, G.~Jin, J.~Chen, J.~Li, R.~Huang, Z.~Chen, Clustering federated
  learning for bearing fault diagnosis in aerospace applications with a
  self-attention mechanism, Aerospace 9~(9) (2022).
\newblock \href {https://doi.org/10.3390/aerospace9090516}
  {\path{doi:10.3390/aerospace9090516}}.

\bibitem{liu2023intelligent}
S.~Liu, H.~Jiang, Z.~Wu, Z.~Yi, R.~Wang, \color{black} intelligent fault
  diagnosis of rotating machinery using a multi-source domain adaptation
  network with adversarial discrepancy matching, Reliability Engineering \&
  System Safety 231 (2023 \color{black}) 109036.

\bibitem{zhang2021conditional}
Q.~Zhang, Z.~Zhao, X.~Zhang, Y.~Liu, C.~Sun, M.~Li, S.~Wang, X.~Chen,
  Conditional adversarial domain generalization with a single discriminator for
  bearing fault diagnosis, IEEE Transactions on Instrumentation and Measurement
  70 (2021) 1--15.

\bibitem{yu2020conditional}
X.~Yu, Z.~Zhao, X.~Zhang, C.~Sun, B.~Gong, R.~Yan, X.~Chen, Conditional
  adversarial domain adaptation with discrimination embedding for locomotive
  fault diagnosis, IEEE Transactions on Instrumentation and Measurement 70
  (2020) 1--12.

\bibitem{li2023remaining}
Y.~Li, Y.~Chen, Z.~Hu, H.~Zhang, \color{black} remaining useful life prediction
  of aero-engine enabled by fusing knowledge and deep learning models,
  Reliability Engineering \& System Safety 229 (2023\color{black}) 108869.

\bibitem{xiong2023adaptive}
J.~Xiong, J.~Zhou, Y.~Ma, F.~Zhang, C.~Lin, \color{black} adaptive deep
  learning-based remaining useful life prediction framework for systems with
  multiple failure patterns, Reliability Engineering \& System Safety 235
  (2023\color{black}) 109244.

\bibitem{cheng2022two}
H.~Cheng, X.~Kong, Q.~Wang, H.~Ma, S.~Yang, \color{black} the two-stage rul
  prediction across operation conditions using deep transfer learning and
  insufficient degradation data, Reliability Engineering \& System Safety 225
  (2022 \color{black}) 108581.

\bibitem{zhang2018transfer}
A.~Zhang, H.~Wang, S.~Li, Y.~Cui, Z.~Liu, G.~Yang, J.~Hu, Transfer learning
  with deep recurrent neural networks for remaining useful life estimation,
  Applied Sciences 8~(12) (2018) 2416.

\bibitem{ding2021remaining}
Y.~Ding, M.~Jia, Q.~Miao, P.~Huang, \color{black} remaining useful life
  estimation using deep metric transfer learning for kernel regression,
  Reliability Engineering \& System Safety 212 (2021\color{black}) 107583.

\bibitem{hu2022remaining}
T.~Hu, Y.~Guo, L.~Gu, Y.~Zhou, Z.~Zhang, Z.~Zhou, Remaining useful life
  estimation of bearings under different working conditions via wasserstein
  distance-based weighted domain adaptation, Reliability Engineering \& System
  Safety 224 (2022) 108526.

\bibitem{zhu2020new}
J.~Zhu, N.~Chen, C.~Shen, A new data-driven transferable remaining useful life
  prediction approach for bearing under different working conditions,
  Mechanical Systems and Signal Processing 139 (2020) 106602.

\bibitem{frederick2007user}
D.~K. Frederick, J.~A. DeCastro, J.~S. Litt, User's guide for the commercial
  modular aero-propulsion system simulation (c-mapss), Tech. rep. (2007).

\bibitem{saxena2008damage}
A.~Saxena, K.~Goebel, D.~Simon, N.~Eklund, Damage propagation modeling for
  aircraft engine run-to-failure simulation, in: 2008 international conference
  on prognostics and health management, IEEE, 2008, pp. 1--9.

\bibitem{li2018remaining}
X.~Li, Q.~Ding, J.-Q. Sun, Remaining useful life estimation in prognostics
  using deep convolution neural networks, Reliability Engineering \& System
  Safety 172 (2018) 1--11.

\bibitem{chao2022fusing}
M.~A. Chao, C.~Kulkarni, K.~Goebel, O.~Fink, Fusing physics-based and deep
  learning models for prognostics, Reliability Engineering \& System Safety 217
  (2022) 107961.

\bibitem{fan2020transfer}
Y.~Fan, S.~Nowaczyk, T.~R{\"o}gnvaldsson, Transfer learning for remaining
  useful life prediction based on consensus self-organizing models, Reliability
  Engineering \& System Safety 203 (2020) 107098.

\bibitem{DBLP:journals/corr/LiWSLH16}
Y.~Li, N.~Wang, J.~Shi, J.~Liu, X.~Hou, Revisiting batch normalization for
  practical domain adaptation, CoRR abs/1603.04779 (2016).
\newblock \href {http://arxiv.org/abs/1603.04779} {\path{arXiv:1603.04779}}.

\bibitem{glorot2010understanding}
X.~Glorot, Y.~Bengio, Understanding the difficulty of training deep feedforward
  neural networks, in: Proceedings of the thirteenth international conference
  on artificial intelligence and statistics, JMLR Workshop and Conference
  Proceedings, 2010, pp. 249--256.

\bibitem{ben2010theory}
S.~Ben-David, J.~Blitzer, K.~Crammer, A.~Kulesza, F.~Pereira, J.~W. Vaughan, A
  theory of learning from different domains, Machine learning 79~(1) (2010)
  151--175.

\end{thebibliography}





\end{sloppypar}
\end{document}